%% file: acl_latex.tex
\pdfoutput=1

\documentclass[11pt]{article}

\usepackage[final]{acl}

\usepackage{times}
\usepackage{latexsym}

\usepackage[T1]{fontenc}

\usepackage[utf8]{inputenc}

\usepackage{microtype}

\usepackage{inconsolata}
\usepackage{amsmath} 
\usepackage{graphicx}
\usepackage{subfigure}
\usepackage{enumerate}
\usepackage{amsfonts} 
\usepackage{amssymb}

\usepackage{listings}
\usepackage{color}

\definecolor{dkgreen}{rgb}{0,0.6,0}
\definecolor{gray}{rgb}{0.5,0.5,0.5}
\definecolor{mauve}{rgb}{0.58,0,0.82}
\title{Understanding and Patching Compositional Reasoning in LLMs}

\author{Zhaoyi Li$^{1,2}$, Gangwei Jiang$^{1,2}$, Hong Xie$^{1}$, Linqi Song$^{2,3}$\textsuperscript{$*$}, Defu Lian$^{1}$\textsuperscript{$*$}, Ying Wei$^{4}\thanks{~~Corresponding authors.}$ \\
$^{1}$School of Computer Science and Technology, University of Science and Technology of China\\
$^{2}$Department of Computer Science, City University of Hong Kong\\
$^{3}$City University of Hong Kong Shenzhen Research Institute \\
$^{4}$School of Computer Science and Engineering, Nanyang Technological University\\
\texttt{\{lizhaoyi777,gwjiang\}@mail.ustc.edu.cn, \{hongx87,liandefu\}@ustc.edu.cn,}\\ \texttt{linqi.song@cityu.edu.hk, ying.wei@ntu.edu.sg}
}
\begin{document}
\maketitle
\input{Working_Draft/sec1_introduction}
\input{Working_Draft/sec2_background}
\input{Working_Draft/sec4_compositional_reasoning_errors}
\input{Working_Draft/sec3_inspecting_and_causal_intervention}
\input{Working_Draft/sec5_locating}
\input{Working_Draft/sec6_CREME}

\input{Working_Draft/sec7_related_work}
\input{Working_Draft/sec8_conclusion}
\bibliography{acl_latex}

\appendix
\input{Working_Draft/Appendix}


\end{document}

%% file: Working_Draft/sec1_introduction.tex
\begin{abstract}
LLMs have marked a revolutonary shift, yet they falter when faced with compositional reasoning tasks.
Our research embarks on a quest to uncover the root causes of compositional reasoning failures of LLMs, uncovering that most of them stem from the improperly generated or leveraged implicit reasoning results.
Inspired by our empirical findings, we resort to Logit Lens and an intervention experiment to dissect the inner hidden states of LLMs. This deep dive reveals that implicit reasoning results indeed surface within middle layers and play a causative role in shaping the final explicit reasoning results.
Our exploration further locates multi-head self-attention (MHSA) modules within these layers, which emerge as the linchpins in accurate generation and leveraing of implicit reasoning results.
Grounded on the above findings, we develop CREME, a lightweight method to patch errors in compositional reasoning via editing the located MHSA modules. 
Our empirical evidence stands testament to CREME's effectiveness, paving the way for autonomously and continuously enhancing compositional reasoning capabilities in language models.

\end{abstract}
\section{Introduction}
Compositional reasoning stands as a pivotal mechanism, 
unlocking the ability of learning systems to decompose complex tasks into manageable sub-tasks and tackle them step-by-step~\cite{lu2023chameleon,Lake2023}. 
Despite the revolutionary impact of Large Language Models (LLMs) on 
the NLP landscape, they struggle at basic compositional reasoning tasks~\cite{dziri2023faith}. 
This shortcoming is specifically highlighted by~\citet{measuring_emnlp2023}, who
brought attention to the concerning ``\textbf{compositionality gap}” in the realm of question-answering tasks.
It was observed that there is a substantial failure rate of $\sim40\%$ in two-hop compositional queries, 
even when they can successfully answer the individual single-hop queries that make up the two-hop question.
Recent attempts improve the compositional reasoning capabilities of LLMs by making them decompose the question and explicitly output the step-by-step rationale with carefully crafted prompting strategies developed by experts~\cite{cot_nips2022,zhou2023leasttomost}. However, understanding the inherent mechanism of compositional multi-hop reasoning inside LLMs and enabling them to autonomously rectify their compositional reasoning errors and continuously improve over time remain largely under-explored.


\begin{figure}[t]
\vspace{-0.1in}
    \raggedright
    \includegraphics[width=1.0\linewidth, keepaspectratio=true]{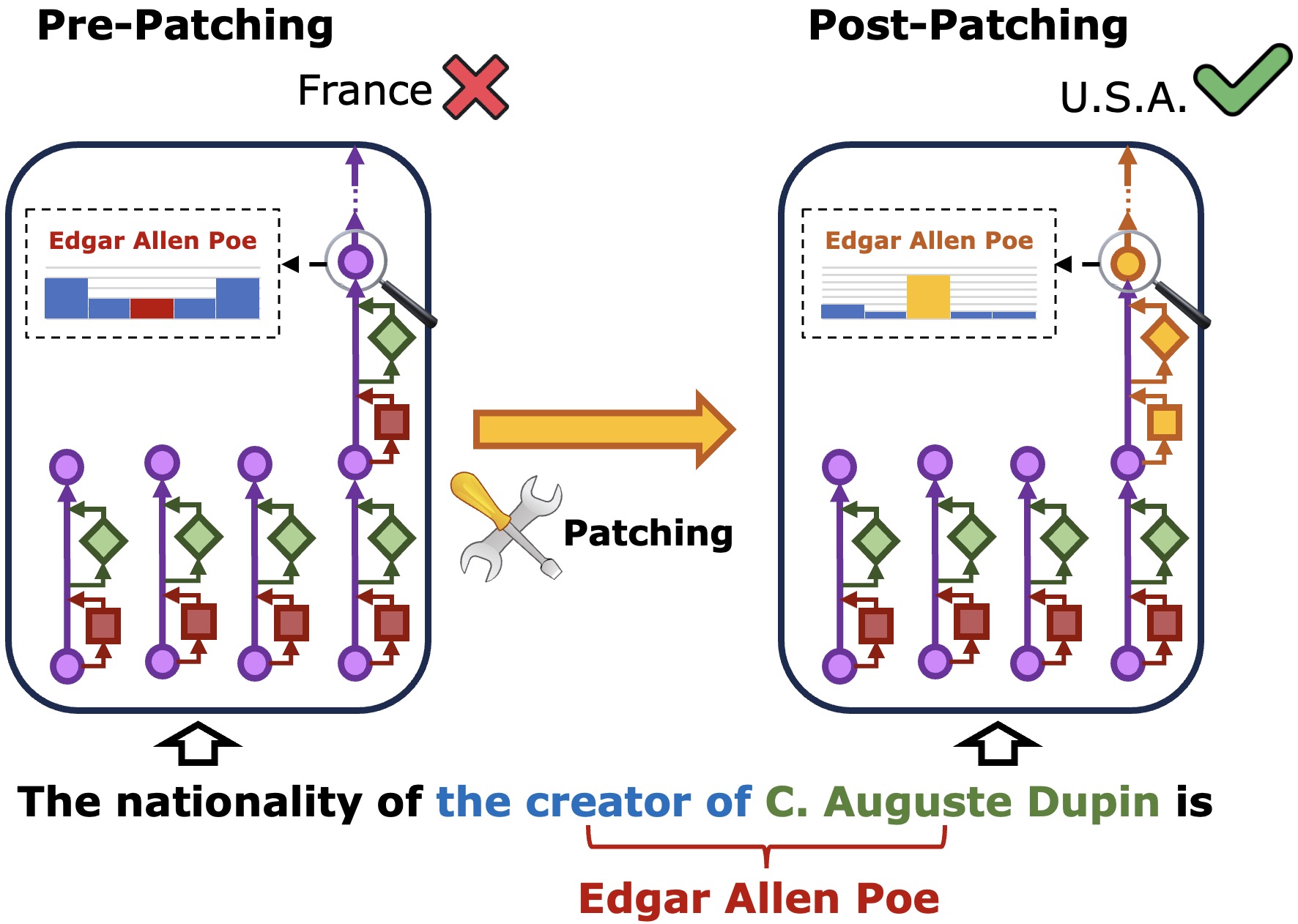}
    \caption{An example of a typical compositional reasoning error pattern :Short-Cut(Section~\ref{sec4:identify error types}). Before patching, LLMs take short-cut reasoning by directly binding “The nationality” and “\textcolor{dkgreen}{C. Auguste Dupin}” (a fictional French detective) to incorrectly predict “France”. After patching, LLMs tend to firstly bind “\textcolor{blue}{the creator of}” and “\textcolor{dkgreen}{C. Auguste Dupin}” to generate “\textcolor{red}{Edgar Allen Poe}” (implicit reasoning result) and then correctly predict “U.S.A.” (explicit reasoning result).}
    \label{fig:intro_fig1}
\vspace{-0.1in}
\end{figure}

This work, therefore, sets out to \emph{firstly} delve into the specific failures to understand (\textbf{RQ1}) what accounts for these compositional reasoning failures and (\textbf{RQ2}) which parts of the LLMs are responsible for them, and \emph{secondly} develop strategies for patching these failures. 
Our initial step involves an analysis of a very recent dataset comprising compositional two-hop knowledge queries~\cite{zhong-etal-2023-mquake},  selectively examining the cases where LLMs fail despite successfully answering the constituent single-hop queries. To ensure our findings and methodologies offer broad applicability, our analyses utilize two widely-used open-sourced LLMs: OpenAlpaca-3B~\cite{openalpaca} and LLaMA-2-7B~\cite{touvron2023llama}. 
Through meticulous examination of the failure instances, we identify three prevalent types of errors.
Utilizing the Logit Lens tool~\cite{LogitLens2020}, each error type highlights a critical shortfall in generating or leveraging the \textbf{implicit reasoning result} necessary for the \textbf{explicit reasoning result}\footnote{Compositional two-hop queries require two-hop reasoning: \textbf{implicit reasoning result} is the first-hop reasoning result; \textbf{explicit reasoning result} is the second-hop reasoning result.}.
This gap is particularly concerning as it contrasts sharply with the intuitive two-hop reasoning process inherent to human cognition. An illustrative example of compositional reasoning error is depicted in Figure~\ref{fig:intro_fig1}, where the model incorrectly concludes its reasoning without properly generating and incorporating the implicit reasoning result. 


The above observations motivate our further empirical inquiry to answer the first question of what accounts for these failures, from the perspective of 
\emph{whether LLMs are indeed aware of implicit reasoning results during compositional reasoning.}
We inspect inner hidden states of LLMs via Logit Lens, from which we observe that implicit reasoning results not only manifest within the LLMs' intermediate layers but also tend to precede the generation of explicit reasoning results, often emerging statistically earlier.
Building on this, we further explore the relationship between implicit and explicit reasoning results through an Intervention~\cite{causal_pearl_2001,li2023emergent} experiment, providing compelling evidence that the emergence of implicit reasoning results within LLMs plays a \textbf{causative role} in the generation of explicit reasoning results.


The next question is, regarding \textbf{RQ2}, 
\emph{in which modules LLMs generate implicit reasoning results?}
Leveraging causal mediation analysis~\cite{meng2022locating,stolfo-etal-2023-mechanistic}, we present both a compositional query and its corresponding second-hop query to the LLM, resulting in the generation of two distinct computation graphs. We then intervene the computation graph $\mathcal{G}_1$, associated with the compositional query, by replacing the output of a single module with its counterpart from the second-hop computation graph $\mathcal{G}_2$. By identifying the modules whose replacement results in a significant enhancement in the predictive probability of the explicit reasoning result, we are able to locate several specific outputs from the Multi-Head Self-Attention (MHSA). 
Intriguingly, the layers pinpointed through this approach show a strong correlation with those identified in preceding Intervention experiments. This congruence reinforces the hypothesis that implicit reasoning results are not only present but are actively consolidated and utilized within these specific layers of the LLM.


Grounded on our findings into RQ1 and RQ2, we develop \textbf{CREME} (Correcting \textbf{C}ompositional \textbf{RE}asoning via \textbf{M}odel \textbf{E}diting), a light-weight model-editing method to patch errors in compositional reasoning.
CREME follows~\citet{NEURIPS2021_c46489a2,meng2022locating} by regarding the output matrix of the located MHSA, $W_o^l$, as a linear associative memory. 
To implement CREME, we designate the input to $W_o^l$ in the computation graph $\mathcal{G}_1$ as $k^*$ and the output from $W_o^l$ in $\mathcal{G}_2$ as $v^*$.  We then proceed to insert the pair $(k^*,v^*)$ into $W_o^l$, ensuring that this insertion disrupts existing memories within $W_o^l$ as minimally as possible. This objective is achieved by solving a convex optimization problem, which strikes a nuanced balance between the integration of new corrective information and the preservation of existing knowledge.

Our main contributions and takeaways are summarized below:
(1) successful compositional reasoning within LLMs hinges on its awareness of generating and leveraging implicit reasoning results;
(2) MHSA modules in the middle layers (18/19-th layer) are significantly in charge of properly generating and leveraging implicit reasoning results;
(3) by leveraging the second-hop computation graph as a reference for editing the located MHSA modules, CREME proves to be highly performing, on correctly answering not only the \emph{query used for editing} $W_o^l$ but also the \emph{paraphrased queries} and \emph{other compositional queries} sharing the first-hop knowledge as well as maintaining little effect on \emph{irrelevant queries}\footnote{Implementation will be available at \url{https://github.com/Zhaoyi-Li21/creme}.}.

%% file: Working_Draft/sec2_background.tex
\section{Background \& Notation}
\begin{table*}[t]
\centering
\resizebox{\linewidth}{!}{
\begin{tabular}{|c|c|c|c|c|c|}
\hline
\textbf{Error type} & \textbf{Input} & \textbf{Implicit result} & \textbf{Correct final result} & \textbf{Predicted final result} & \textbf{Proportion} \\
\hline
Distortion & \textcolor{blue}{The nationality of} \textcolor{red}{the performer of the song ``I Feel Love”} is & \textcolor{red}{Donna Summer}  & \textcolor{blue}{United States of America} & United Kingdom $\backslash$ Italy & $15\%$\\
\hline
Incomplete Reasoning & \textcolor{blue}{The head of state of} \textcolor{red}{the country where ORLAN holds citizenship} is & \textcolor{red}{France} & \textcolor{blue}{Emmanuel Macron} & France & $36\%$\\
\hline
Hasty Answer \uppercase\expandafter{\romannumeral1} & \textcolor{blue}{The capital city of} \textcolor{red}{the country where ``Work from Home” originated} is & \textcolor{red}{United States of America} & \textcolor{blue}{Washington, D.C.} & Los Angeles $\backslash$ New York& $14\%$\\
\hline
Hasty Answer \uppercase\expandafter{\romannumeral2} & \textcolor{blue}{The home country of} \textcolor{red}{the sport associated with Giorgio Chinaglia} is & \textcolor{red}{association football} & \textcolor{blue}{England} & Italy & $12\%$\\
\hline
\end{tabular}
}
\vspace{-0.1in}
\caption{
Specific examples in $D_{gap}$ for three types of common errors. 
``Predicted final result” column refers to the wrong answers output by LLaMA-2-7B. For each compositional knowledge, we query the language model with at most three paraphrased questions, and hence the predicted answers can be multiple. In the \textbf{Proportion} column, we calculate the proportions for each type of errors observed in the 200 two-hop examples from MQuAKE dataset~\cite{zhong-etal-2023-mquake}. 
}
\vspace{-0.1in}
\label{tab:error_type}
\end{table*}
\subsection{Logit Lens}
\label{sec:logit_lens}
Logit Lens~\cite{LogitLens2020} is a widely used for inspecting hidden states of LLMs~\cite{dar-etal-2023-analyzing, geva-etal-2023-dissecting, katz-belinkov-2023-visit, memoryinjections_blackboxnlp2023}. 
The key idea of Logit Lens is thus to interpret hidden states in middle layers of LLMs via projecting them into the output vocabulary space with the \textbf{LM head} $W_u$. When presented with a specific hidden state $h_l^t$ and a set of target tokens $T_{tgt}$, 
the Logit Lens is given as follows: 
\begin{align}
\label{eq:logit_lens}
    & L(h_l^t, T_{tgt}) = \frac{1}{|T_{tgt}|} \sum_{k\in T_{tgt}} p_l^t[k], \\
    & p_l^t = \text{softmax}(v_l^t) = \text{softmax}(h_l^t W_u),
\label{eq:prob}
\end{align}
where 
$L(h_l^t, T_{tgt})$ measures how much information around $T_{tgt}$ is contained in $h_l^t$. Note that typically there are multiple tokens in $T_{tgt}$ and we separately calculate the probabilities for all of these tokens and adopt their mean value to represent the whole $T_{tgt}$, similar with ~\cite{chanin2024identifying}. There are other works~\cite{memoryinjections_blackboxnlp2023,yang2024large} use the first token in $T_{tgt}$ to represent it, which sometimes can bring loss of information in $T_{tgt}$ and additional conflicts when processing targets like “United Kingdom” and “United States” as well.
\subsection{Compositional Reasoning and Dataset}
\label{sec2:comp_reasoning}
Compositional knowledge refers to knowledge items that are the compositions of several single-hop sub-knowledge items. 
Compositional reasoning refers to the ability to answer the queries on compositional knowledge (e.g., verbalized in format of QA or Cloze-Test) via a \textbf{step-by-step reasoning} process.
We denote a single-hop knowledge as a triple $(s, r, o)$, where $s,r,o$ represents subject, relationship and object respectively.
The composed compositional two-hop knowledge is denoted as $(s_1,r_1,o_1)\oplus (s_2,r_2,o_2)$ where subscripts $1$ and $2$ represent the \textbf{first-hop} and \textbf{second-hop} sub-knowledge (requiring $o_1 = s_2$ so that they can compose together).
The dataset $\mathcal{D}$ (Appendix~\ref{appendix:datasets}) we used in this paper is sourced from ~\citet{zhong-etal-2023-mquake}. For each datum in $\mathcal{D}$, it contains: (1) the compositional query on the compositional knowledge $(s_1,r_1,o_1)\oplus (s_2,r_2,o_2)$, (2) the first-hop query on $(s_1,r_1,o_1)$, (3) the second-hop query on $(s_2,r_2,o_2)$, and (4) the \textbf{implicit reasoning result} $o_1$ and the \textbf{explicit reasoning result} $o_2$. 
By way of example, the first-hop query is ``What is the sport associated with ($r_1$) Giorgio Chinaglia ($s_1$)? \underline{association football} ($o_1$)”, the second-hop query is ``What is the home country of ($r_2$) association football ($s_2$)? \underline{England} ($o_2$)” and the compositional query can be verbalized as ``What is the home country of ($r_2$) the sport associated with ($r_1$) Giorgio Chinaglia ($s_1$)? \underline{England} ($o_2$)”.

%% file: Working_Draft/sec4_compositional_reasoning_errors.tex
\section{Analyzing Compositional Reasoning Errors}
\label{sec4:inference and errors}
\label{sec:infer_dataset}
\label{sec4:identify error types}
\label{sec4:explanation to errors}
Grounded on the observation of ~\citet{measuring_emnlp2023}, we dive into the compositional reasoning failures: 
we identify three types of common errors among such failures and attribute the cause of these common errors to the failure of generating implicit reasoning result properly via inspecting hidden states.
\paragraph{Three types of Common Errors}
We query LLMs with all of compositional queries and the corresponding single-hop queries in $\mathcal{D}$. 
We filter out two subsets of $\mathcal{D}$: $\mathcal{D}_{single}$ and $\mathcal{D}_{gap}$.
For each datum $(s_1,r_1,o_1)\oplus (s_2,r_2,o_2)$ in $\mathcal{D}$, $\mathcal{D}_{single}$ contains the datum where the both of $(s_1, r_1, o_1)$ and $(s_2, r_2, o_2)$ are successfully answered. Among $\mathcal{D}_{single}$, $\mathcal{D}_{gap}$ contains the datum where the answer for the compositional queries $(s_1,r_1,o_1)\oplus (s_2,r_2,o_2)$ are mis-predicted.\footnote{Please find details in Appendix~\ref{appendix:inference_prompt}.}
In our analysis of $\mathcal{D}_{gap}$, we have discerned a few common patterns shared among a substantial portion of the failures. Consequently, we have delineated three predominant types of errors, each characterized by distinct features, as outlined below.
\textbf{Distortion}: LLMs fail to effectively generate implicit reasoning results in the reasoning process. The predicted answer for 
the first example in Table~\ref{tab:error_type} 
is either United Kingdom or Italy. Considering both as countries (corresponding to nationality ($r_2$)), we conclude that
the information about Donna Summer ($o_1$) distorts in middle hidden states. 
\textbf{Incomplete Reasoning}: LLMs directly output the first-hop reasoning result ($o_1$). In the second example of Table~\ref{tab:error_type}, LLaMA-2 outputs France ($o_1$) while the correct answer requires further reasoning. the head of state of ($r_2$) France ($o_1$) is Emmanuel Macron ($o_2$).
\textbf{Hasty Answer}: 
LLMs predict the result without carefully reasoning. We further subdivide this type of errors into two categories:
\textbf{\uppercase\expandafter{\romannumeral1}}: LLMs finally predict a close result based on the implicit reasoning result. For the third example in Table~\ref{tab:error_type}: LLMs predict Los Angeles or New York, both of which are famous city in the U.S.A., implying that LLMs manage to generate the implicit result ($o_1$:U.S.A.) while fails to incorporate ``the capital of” ($r_2$) to generate final result $o_2$.
\textbf{\uppercase\expandafter{\romannumeral2}}: LLMs take short-cut instead of step-by-step reasoning, leading to incorrect answers.
Consider the fourth example in Table~\ref{tab:error_type}: the correct reasoning process should be (1): the sport associated with ($r_1$) Giorgio Chinaglia ($s_1$) is \underline{association football} ($o_1$); followed by (2): the home country of ($r_2$) association football ($o_1$) is \underline{England} ($o_2$).
However, LLMs erroneously attribute Italy as the answer. 
This misstep is attributed to LLMs' tendency to directly associate Giorgio Chinaglia ($s_1$) -- noted for his Italian nationality -- with the home country of the sport ($r_2$). We calculate the proportions for each type of errors observed in the 200 two-hop examples of MQuAKE dataset~\cite{zhong-etal-2023-mquake} and post the statistics in Table~\ref{tab:error_type} to demonstrate the commonness of these error types.
\paragraph{Analysis and Possible Explanation}
\begin{figure*}
    \centering
    \subfigure[\scriptsize{\textbf{Distortion:Comp}
    }
    ]{\includegraphics[width=0.24\linewidth]{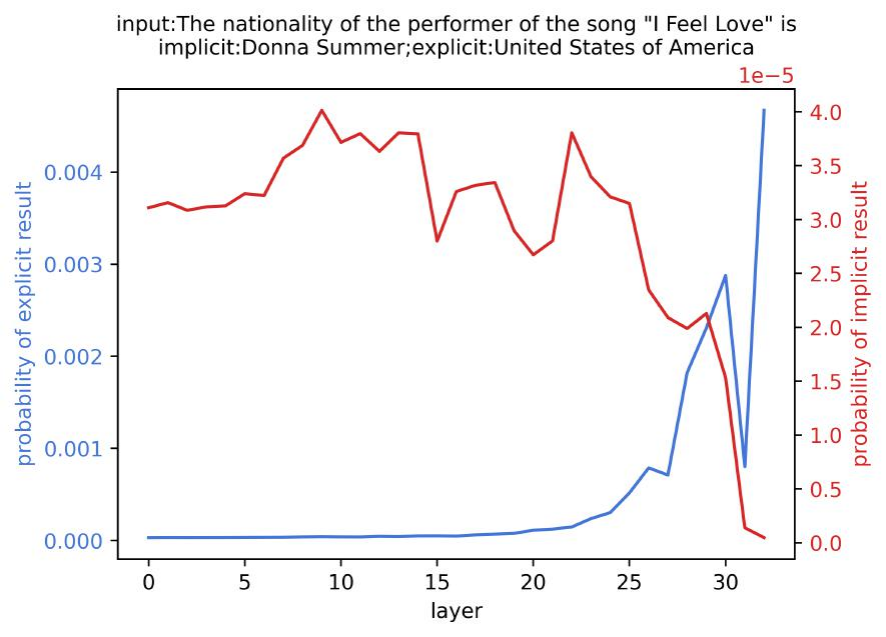}} 
    \subfigure[\scriptsize{\textbf{Incomplete Reasoning:Comp}
    }]{\includegraphics[width=0.24\linewidth]{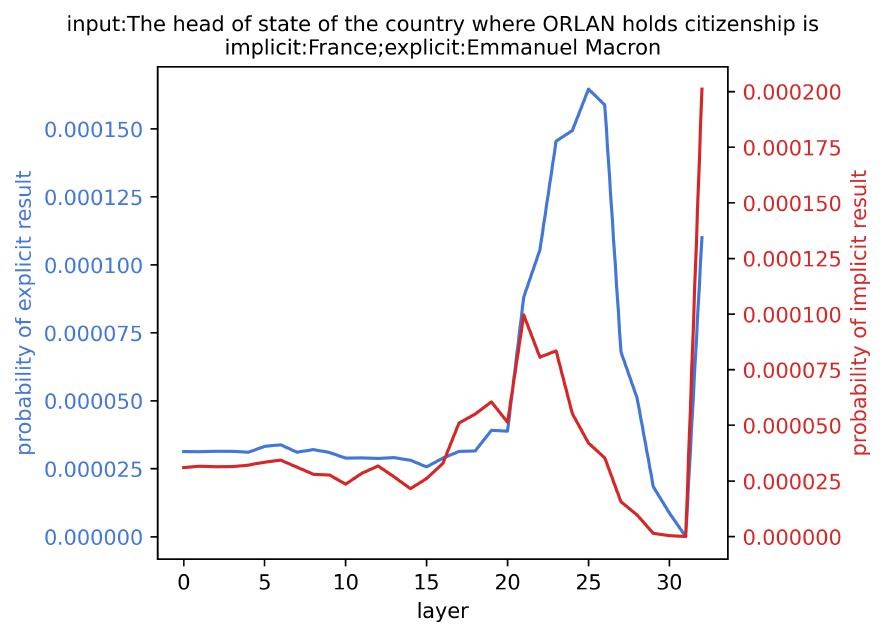}}
    \subfigure[\scriptsize{\textbf{Hasty Answer \textbf{\uppercase\expandafter{\romannumeral1}}:Comp}
    }]{\includegraphics[width=0.24\linewidth]{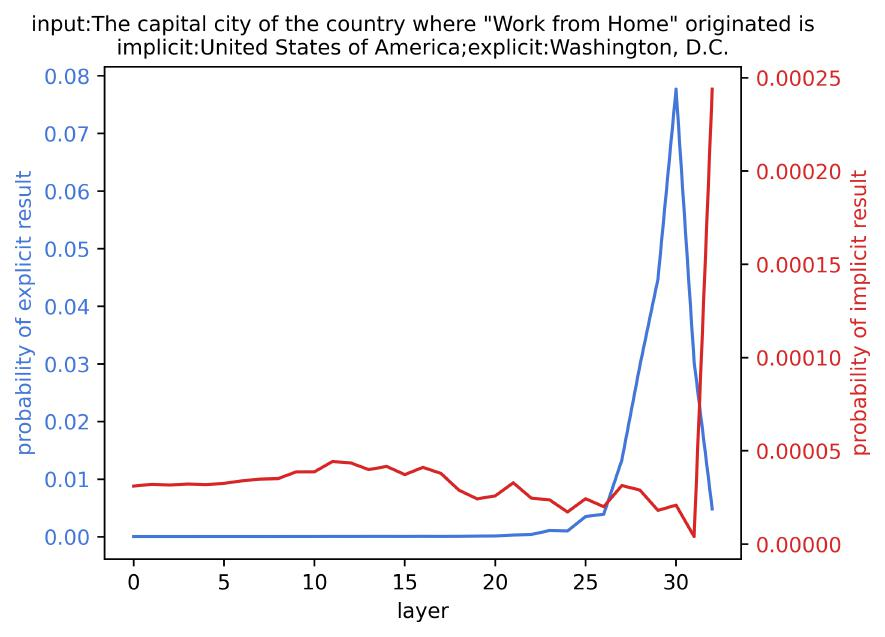}}
    \subfigure[\scriptsize{\textbf{Hasty Answer \textbf{\uppercase\expandafter{\romannumeral2}}:Comp}
    }]{\includegraphics[width=0.24\linewidth]{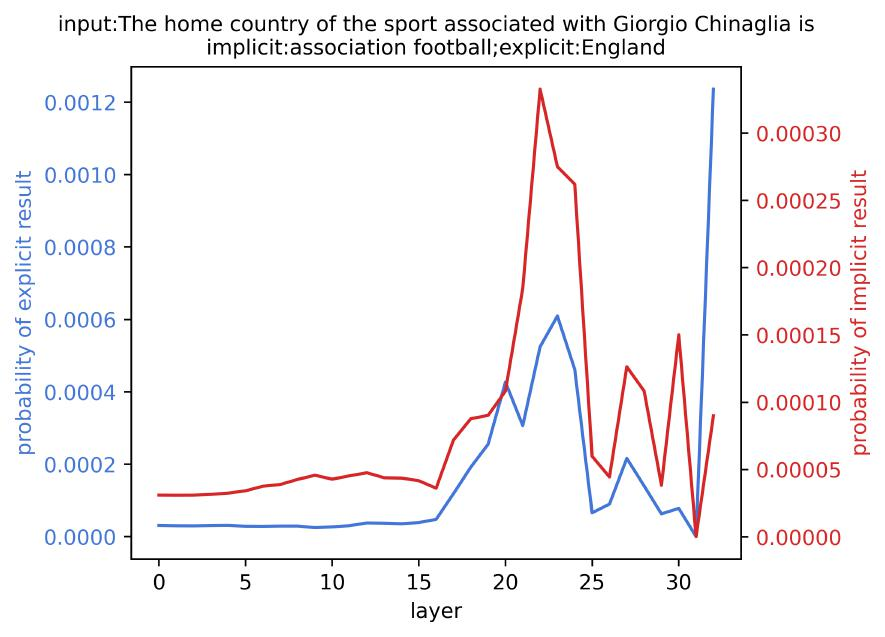}} 
    \subfigure[\scriptsize{\textbf{Distortion:Reference}
    }]{\includegraphics[width=0.24\linewidth]{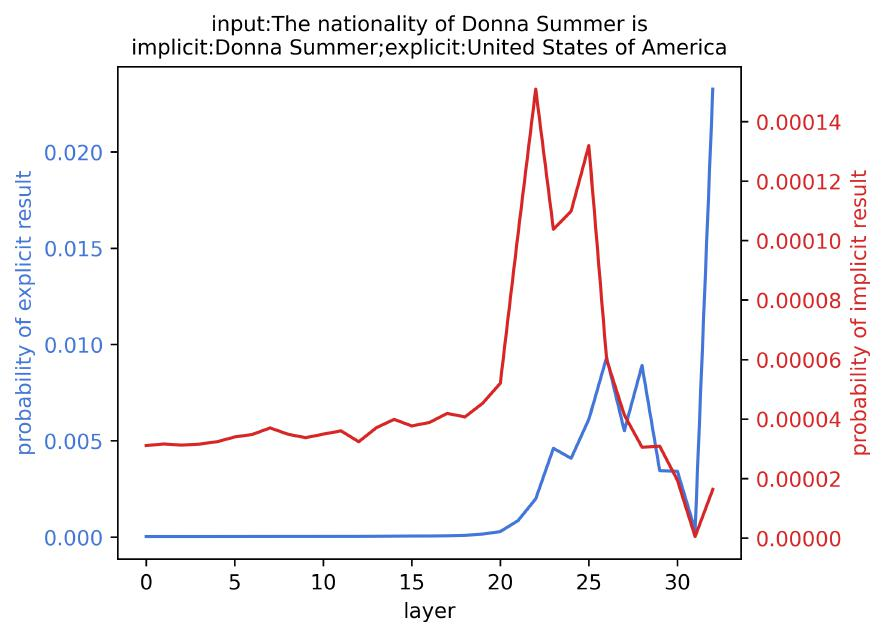}} 
    \subfigure[\scriptsize{\textbf{Incomplete Reasoning:Reference}
    }]{\includegraphics[width=0.24\linewidth]{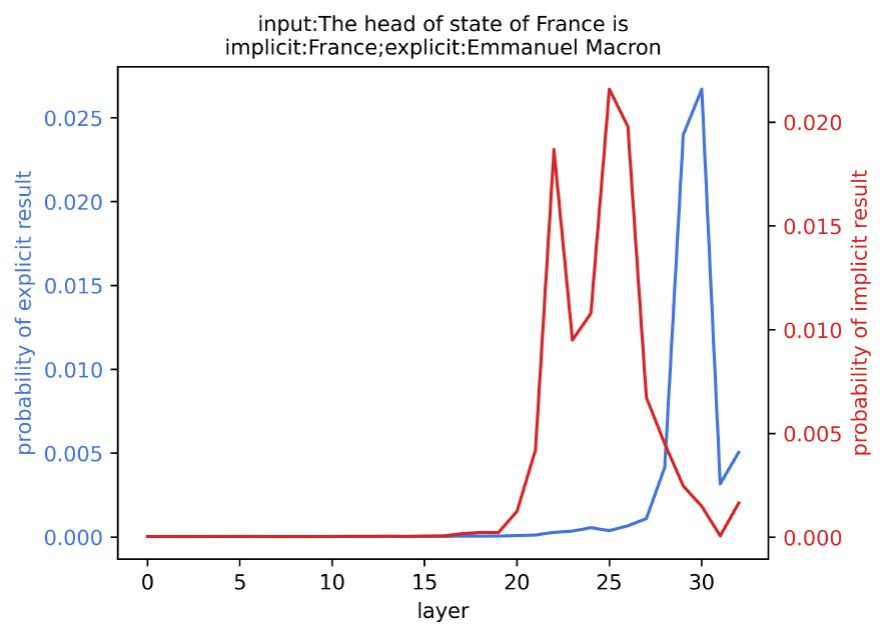}}
    \subfigure[\scriptsize{\textbf{Hasty Answer \textbf{\uppercase\expandafter{\romannumeral1}}:Reference}
    }]{\includegraphics[width=0.24\linewidth]{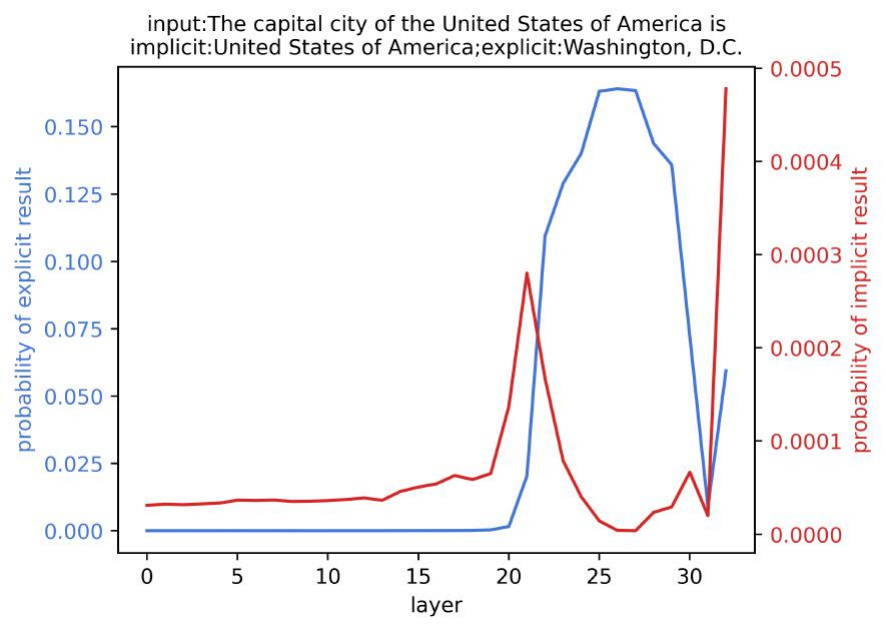}}
    \subfigure[\scriptsize{\textbf{Hasty Answer \textbf{\uppercase\expandafter{\romannumeral2}}:Reference}
    }]{\includegraphics[width=0.24\linewidth]{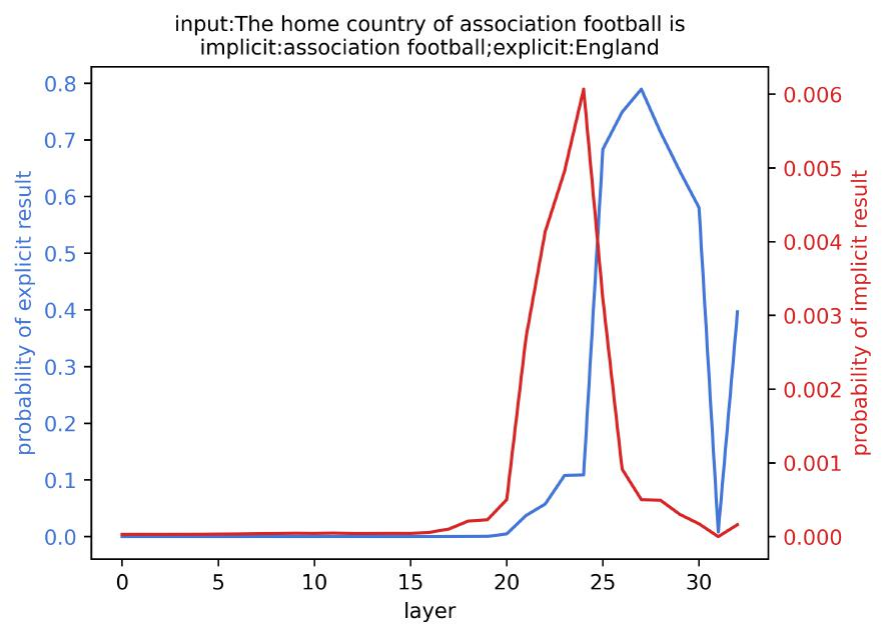}} 
\vspace{-0.1in}
    \caption{Logit Lens results of examples of three error types. \textbf{Comp} is the result for compositional two-hop query; \textbf{Reference} is the result for the corresponding second-hop query (as the reference for the compositional query). \textcolor{red}{red} and \textcolor{blue}{blue} lines trace the \textcolor{red}{implicit} and \textcolor{blue}{explicit} results respectively. y-axis represents the inspecting value (Eqn.~\ref{eq:logit_lens}).}
    \label{fig:error_logit_lens}
\vspace{-0.1in}
\end{figure*}
We aim to analyze the cause of these errors via inspecting the inner workings of LLMs. 
We depict Logit Lens results of the examples of Table~\ref{tab:error_type} (compositional queries) and their references (corresponding second-hop queries) in Figure~\ref{fig:error_logit_lens}, Leveraging Eqn.~\ref{eq:logit_lens}.
Note that in Figure~\ref{fig:error_logit_lens}, results of second-hop inputs (subfigure (e)$\sim$(h)) align well with the results in Figure~\ref{fig:logit_lens_init_observation}. 
However, when we set our sights on results of compositional inputs (subfigure (a)$\sim$(d)), we get clues about the above three error types. 
In (a, \textbf{Distortion}) we observe that the peak for $o_1$ does not emerge at all (probability$\sim\frac{1}{|V|}$), implying the distortion of the predictive information for $o_1$ 
by context.
In (b, \textbf{Incomplete Reasoning}), though $o_1$ emerge in middle layers, it is not intense enough (in comparison with (f)) to arise the final result $o_2$. In Figure~\ref{fig:error_logit_lens_incomplete_reasoning}, we show another example where the peak probability of $o_1$ aligns well with the result of the reference and correctly predict $o_2$.
In (c, \textbf{Hasty Answer \uppercase\expandafter{\romannumeral1}}) we observe that $o_1$ emerge at the last layer, which is too late to incorporate second-hop information to generate $o_2$.
In (d, \textbf{Hasty Answer \uppercase\expandafter{\romannumeral2}}) although $o_1$ (association football) also emerges, the peak probability of $o_1$ is much lower than its reference (h). For comparison, we plot the Logit Lens of ``the home country ($r_2$) of Giorgio Chinaglia ($s_1$)'' for ``Italy'' 
in Figure~\ref{fig:error_logit_lens_short_cut}, which aligns with its corresponding compositional query well,
advocating that LLMs predict through short-cut.
In summary, all of these errors can be attributed to improperly generating implicit reasoning results. The implicit reasoning results either (1): do not notably emerge (\textbf{Distortion}) or (2): emerge but not intensely or timely enough to raise the explicit reasoning results(\textbf{Incomplete Reasoning} and \textbf{Hasty Answer}).

%% file: Working_Draft/sec3_inspecting_and_causal_intervention.tex
\section{Analyzing the Inner Hidden States of LLMs for Compositional Reasoning}
\label{sec3:analyzing compositional reasoning}
\begin{figure}
    \centering
    \subfigure[compositional queries]{\includegraphics[width=0.48\linewidth]{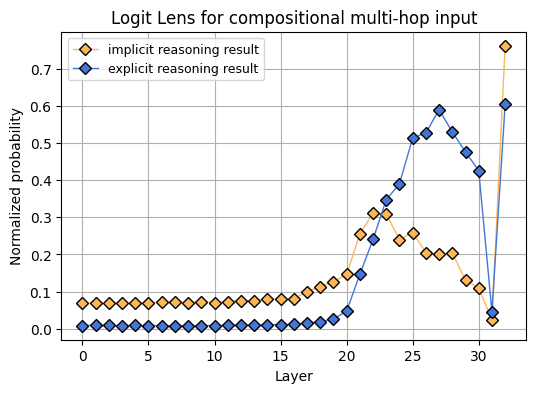}} 
    \subfigure[the second-hop queries]{\includegraphics[width=0.48\linewidth]{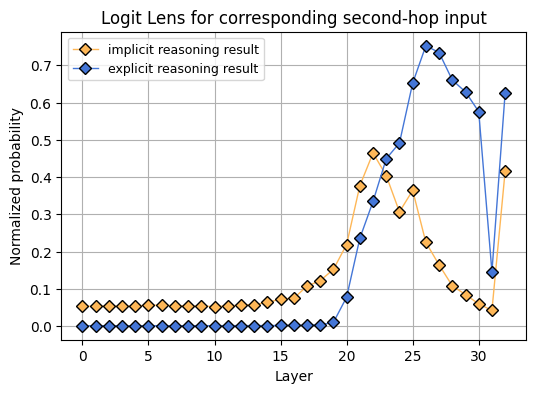}} 
    \vspace{-0.1in}
    \caption{Logit Lens inspecting results with LLaMA-2-7B. (a) refers to the averaged result for inputs of compositional two-hop queries and (b) refers to the averaged result for second-hop queries. x-axis refers to the layer; y-axis refers averaged Logit Lens values after min-max normalization (i.e., the original values are linearly mapped to $[0,1]$). Yellow line and blue line refers to implicit results and explicit results respectively.}
    \label{fig:logit_lens_init_observation}
\vspace{-0.1in}
\end{figure}

Providing that LLMs are capable to perform compositional step-by-step reasoning~\cite{hou-etal-2023-towards}, we hypothesize that they generate the implicit reasoning result $o_1$ (the notation is aligned with Section~\ref{sec2:comp_reasoning}) in the process of compositional reasoning, before finally obtaining the explicit reasoning result $o_2$. 
We inspect inner hidden states of LLMs via Logit Lens (Section~\ref{sec3:inspecting}) and observe that implicit reasoning results emerge in middle layers, implying that they may play a role in the compositional reasoning process (Section~\ref{sec3:init observation}). To verify this hypothesis, we design an intervention experiment (Section~\ref{sec3:causal intervention}) and demonstrate the emerging of $o_1$ has causal effect on predicting $o_2$ in the output layer (Section~\ref{sec3:result_conclusion}). 
\subsection{Inspecting hidden states of LLMs}
\label{sec3:inspecting}
\label{sec3:init observation}
Given an input of a compositional two-hop knowledge item $(s_1,r_1,o_1)\oplus (s_2,r_2,o_2)$, we denote $h_l,(l\in[1..L])$ as the hidden states at the position of \textbf{last input token} and $l$-th layer. 
Leveraging Eqn.~\ref{eq:logit_lens} we tokenize implicit result $o_1$ and explicit $o_2$ into tokens: $R_i$ (implicit) and $R_e$ (explicit), and inspect the information about $R_i$ and $R_e$ in $h_l$: $L(h_l,R_i)$ and $L(h_l,R_e)$. 
We present the inspecting results averaging over $\mathcal{D}$ with LLaMA-2-7B in Figure~\ref{fig:logit_lens_init_observation}(a). We observe that (1) both $L(R_i,h_l)$ and $L(R_e, h_l)$ reach a peak and then decline with the layer increasing; (2) the peak of $L(R_i,h_l)$ appears at the earlier layer than $L(R_e, h_l)$. Then we use the corresponding second-hop queries $(s_2, r_2, o_2)$ ($s_2=o_1$) to repeat the inspecting experiment. The averaged result is depicted in Figure~\ref{fig:logit_lens_init_observation}(b). We get the similar observations with the compositional two-hop queries, to some extent aligning their reasoning processes: \emph{both of the compositional query (implicitly containing $o_1$) and the second-hop knowledge query (explicitly containing $o_1$) generate $o_1$ in hidden states of middle layers before generating $o_2$}.

The insights gleaned from the emergence of implicit results suggest a potential influence of them on compositional reasoning. In the subsequent analysis, we endeavor to elucidate \emph{how implicit reasoning results, embedded within the hidden states of intermediary layers, exert a causal impact on the generation of explicit reasoning results}.
\subsection{Verifying the Hypothesis via Intervention}
\label{sec3:causal intervention}
\label{sec3:result_conclusion}
\begin{figure}
    \centering
    \subfigure[LLaMA-2-7B]{\includegraphics[width=0.49\linewidth]{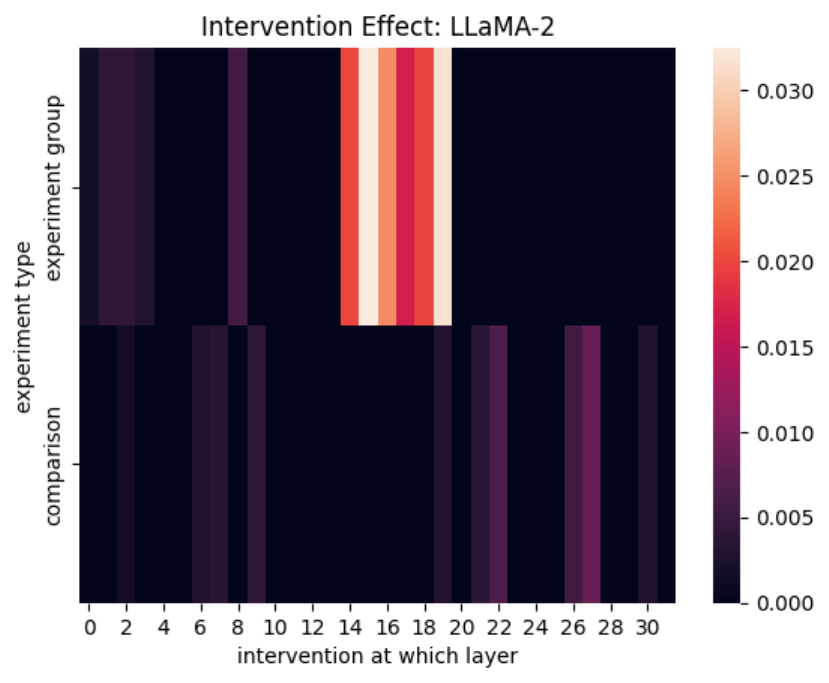}} 
    \subfigure[OpenAlpaca-3B]{\includegraphics[width=0.49\linewidth]{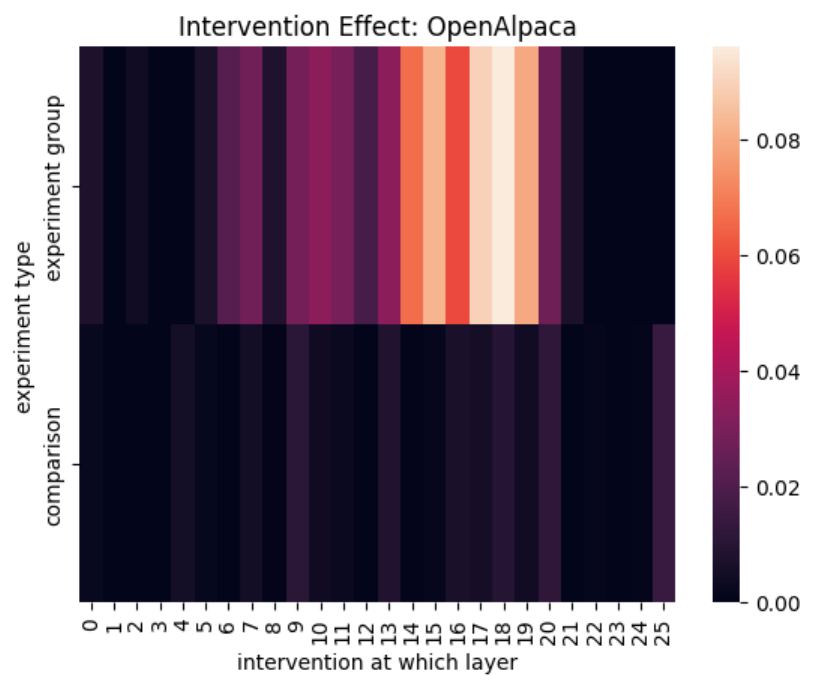}}
    \caption{Intervention experiment: Brighter color indicates the intervention effect is more significant. In each subfigure, the upper row refers to the experiment group and the lower row refers to the comparison group. Note that for better visualization, we clip the effect value ($\le0$) to $0$ for both of the experiment and comparison groups. }
    \label{fig:debias_dataset}
\end{figure}

We recall the notations defined before. 
The tokenizations of $o_1$ and $o_2$ are $R_i$ and $R_e$; the hidden state of the last token at the $l$-th layer is $h_l$. Accordingly, the probability distribution over the output vocabulary set $V$ (with Eqn.~\ref{eq:prob}) is $p_l = \text{softmax}(v_l) =\text{softmax}(h_l\cdot W_u) \in\mathbb{R}^{|V|}$. Our aim is to demonstrate how the information about $o_1$ encoded in hidden states of middle layers plays a causal role in the prediction of 
$o_2$. 
The technique of \textbf{Intervention}~\cite{causal_pearl_2001,li2023emergent} fits the objective, where
we strategically intervene on these inner hidden states to eliminate the information related to $o_1$ (through Logit Lens) and observe the resultant impact on  predicting $o_2$.
\paragraph{Intervention} We define the intervention $\mathcal{I}_l: h_l \rightarrow h_l^{*}$, where $h_l^{*}$ denotes the intervened hidden state. $v_l^{*}$ is the corresponding logits (through Logit Lens) of $h_l^{*}$: $v_l^{*}=h_l^{*}\cdot W_u$.
Denoting that (before intervention) $v_{min} = \min\limits_{0\leq j < |V|} \{v_l[j] \}$, we expect $v_l^{*}$ meets the following constraints:
\begin{equation}
v_l^{*}[j]=\left\{
\begin{aligned}
& v_{min}, & j\in R_i, \\
& v_l[j], & j \in [0..|V|)/R_i, \\
\end{aligned}
\right.
\end{equation}
Which means, observing from Logit Lens, we \textbf{eliminate the bias} on $o_1$ in $h_l^{*}$ in the computation graph and minimize the side effects on the rest tokens\footnote{More discussion please refer to Appendix~\ref{appendix:implment_causal_intervention}.}.
We solve the linear system $v_l^{*} =  h_l^{*} \cdot W_u$ to get $h_l^{*}$: $h_l^{*} =  v_l^{*} W_u^T (W_u W_u^T)^{-1}$ (in case that $W_u W_u^T$ is not full-rank, we use the Moore–Penrose inverse~\cite{bams/1183425340} instead). In our implementation, we calculate the difference value for the purpose of numerical stability:
\begin{equation}
\begin{aligned}
   h_l^{*} = h_l + (v_l^{*}-v_l) W_u^T (W_u W_u^T)^{-1}.
\end{aligned}
\end{equation}
\paragraph{Effect}
We define the effect $\mathcal{E}_l$ of an intervention $\mathcal{I}_l$ is the difference between probabilities of predicting $o_2$ (tokenization: $R_e$) at the output layer $L$ before and after the intervention:
\begin{equation}
    \mathcal{E}_l = p_L[R_e] - p_{L}^{\mathcal{I}_l}[R_e].
\end{equation}
Ideally, we expect the intervention $\mathcal{I}_l$ has the effect of decreasing the probability of predicting the explicit reasoning result $o_2$ (i.e., $\mathcal{E}_l>0$).
\paragraph{Result}
The Intervention experiment results (averaged over $\mathcal{D}$) are depicted in Figure~\ref{fig:debias_dataset}. For each experiment group, we set a \textbf{comparison group} where we intervene on $|R_i|$ tokens that are \textbf{randomly sampled} from $V$. Comparing experiment groups and comparison groups, we observe there exist apparent positive effects ($\mathcal{E}_l>0$) when intervening middle layers (for both LLaMA-2 and OpenAlpaca, positive effects appear in $15\text{-th}\text{ to }20\text{-th}$ layers) for experiment groups, suggesting that the information about $o_1$ may be generated and utilized for generating $o_2$ in these layers.
Meanwhile, there is nearly no notable positive effect for comparison groups across all layers. 
The results verify our hypothesis that the information around implicit reasoning results in middle layers play a role in predicting explicit reasoning results.

%% file: Working_Draft/sec5_locating.tex
\section{Locating Important Modules}
\label{sec5:locating_important_modules}
In previous analysis, we attribute compositional reasoning errors to improperly generating implicit reasoning results. 
In this section, we aim to investigate if there sparsely exist some ``key” modules (i.e., MHSA or MLP)\footnote{We introduce the LLM architecture in Appendix~\ref{appendix:llm_arch}} in LLMs that are responsible for properly generating implicit reasoning results in hidden states of middle layers. 
\subsection{Locating Methodology}
In Section~\ref{sec4:explanation to errors}, we observe that if inspecting results of the compositional query and its corresponding second-hop query align well, the compositional reasoning process is usually in smooth going.
Given this, combining the key idea in Causal Mediation Analysis~\cite{meng2022locating, stolfo-etal-2023-mechanistic}, we propose the following locating method.
(1) We run the LLM twice: once with the compositional query in $\mathcal{D}_{gap}$ in the length of $T_1$ and once with its corresponding second-hop query in the length of $T_2$. For the compositional pass, we denote the module outputs in the computation graph as $\{\eta_l^t|\eta\in\{a,m\},l\in[1..L], t\in[1..T_1]\}$ ($a$ for MHSA, $m$ for MLP, $l$ indexing layers, $t$ indexing tokens). For the second-hop pass, we denote the outputs as $\{\hat{\eta}_l^t|\eta\in\{a,m\}, l\in[1..L], t\in[1..T_2]\}$.
(2) We replace a single module output of interest in the compositional pass computation graph with its counterpart in the second-hop pass computation graph. We focus on two token positions: the \textbf{last subject token} (which refers to $(s_1,r_1)$ for compositional queries, e.g., ``the sports associated with Giorgio Chinaglia”) and the \textbf{last token}\footnote{These two positions have been demonstrated as most informative for factual reasoning~\cite{meng2022locating}.}.
We denote the original probability of predicting $o_2$ as $p(o_2)$ and the probability after replacement as $p(o_2|\hat{\eta}^{t^*}_l \rightarrow \eta^{t}_l)$.
(3): We define the effect of the replacement $\hat{\eta}^{t^*}_l \rightarrow \eta^{t}_l$ as $p(o_2|\hat{\eta}^{t^*}_l \rightarrow \eta^{t}_l)-p(o_2)$.
\subsection{Insight}
\begin{figure}
    \centering
    \subfigure[LLaMA-2-7B]{\includegraphics[width=0.49\linewidth]{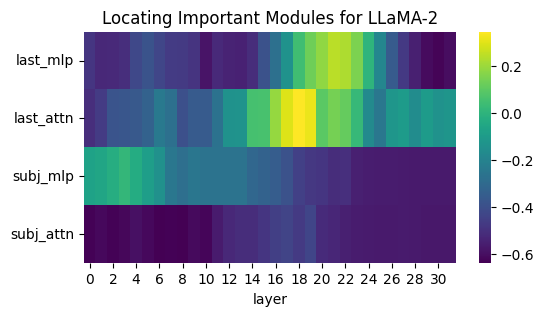}} 
    \subfigure[OpenAlpaca-3B]{\includegraphics[width=0.49\linewidth]{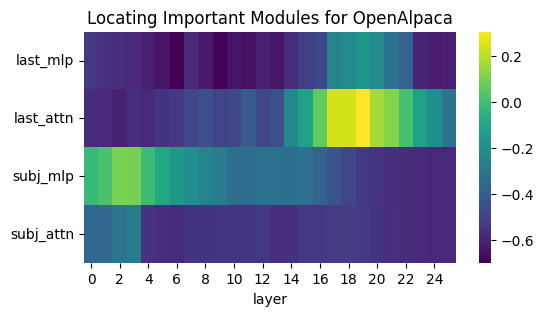}} 
\vspace{-0.1in}
    \caption{AIE for replacements. 
    ``last”: last token; ``subject”: last subject token; ``mlp”: replace the MLP output; ``attn”: replace the MHSA output.
    Brighter positions indicate replacements of larger effect (more important).}
    \label{fig:locating_dataset}
\vspace{-0.1in}
\end{figure}
We depict the \textbf{Average Indirect Effect} (AIE) of replacements over modules, tokens, and layers in Figure~\ref{fig:locating_dataset}. We observe that replacing the MHSA output at the position of (last-token, $18\backslash19$-th layer) has the largest effect on finally predicting the correct answer $o_2$. Interestingly, this coincides with the intervention experiment results in Figure~\ref{fig:debias_dataset}, implying that MHSA modules of these positions play an important role in properly accumulating and leveraging implicit reasoning results.

%% file: Working_Draft/sec6_CREME.tex
\section{Patching Compositional Reasoning}
\label{sec:creme}
Grounded on the empirical insights in Section~\ref{sec3:analyzing compositional reasoning} and Section~\ref{sec5:locating_important_modules}, we are poised to introduce the CREME
approach, designed to  
correct compositional reasoning failures via editing the parameters of MHSA at the \textbf{located positions}. 
We demonstrate its superiority through comparative analyses with two recent baselines for correcting compositional reasoning~\cite{memoryinjections_blackboxnlp2023,ghandeharioun2024patchscopes} and a a widely recognized model editing baseline~\cite{meng2022locating}.

Specifically, our edit objective is the MHSA output matrix at the $l$-th layer $W_O^l$ (for detailed description, please refer to Eqn.~\ref{eq:attention_output}).
Following~\citet{NEURIPS2021_c46489a2}, we view $W_O^l$ as a linear associative memory~\cite{5008975}: $W_O^l\in\mathbb{R}^{d\times d}$ operates as a key-value store for a set of vector keys $K=[k_1|k_2|...]$ and corresponding vector values $V=[v_1|v_2|...]$, by solving $(W_O^l)^TK=V$.

For a given compositional query and its corresponding second-hop query, we run the LLM twice: once with the compositional query and once with the second-hop query.
In the first pass with the compositional query, the \textbf{input} of $W_O^l$ at the last token position is $k_*\in\mathbb{R}^{d\times1}$; in the second pass with the corresponding second-hop query, the \textbf{output} of $W_O^l$ at the last token position is $v_*\in\mathbb{R}^{d\times1}$.
We aim to edit $W_O^l$ to $\hat{W_O^l}$ such that:
\begin{align*}
    \text{minimize }\lVert (\hat{W_O^l})^TK-V \rVert_F^2 \text{ and } (\hat{W_O^l})^Tk_*=v_*,
\end{align*}
where the Frobenius norm guarantees consistent predictions on irrelevant queries while the constraint implements the edit as an insertion of $(k_*, v_*)$ into the linear memory $\hat{W_O^l}$.
Following ~\citet{meng2022locating}, we derive a closed form solution: $\hat{W_O^l}=W_O^l+(C^{-1}k_*)^T\Lambda^T$ where $C=KK^T$ is a constant to estimate the uncentered covariance of $k$ (note that $k$ is randomly sampled from Wikipedia to represent irrelevant queries) and $\Lambda=(v_*-(W_O^l)^Tk_*)/(C^{-1}k_*)^Tk_*$. 
Hopefully, the edited LLMs are able to properly generate implicit reasoning results at the located position and thus alleviate failures of compositional reasoning.

What is worthy noting is that applying CREME only requires \textbf{a single case} of (compositional query, referenced second-hop query) while the patching effect can \textbf{generalize to many other related cases}. We showcase the effect of CREME in Table~\ref{tab:showcase1} and quantitatively discuss this generalization effect in Section~\ref{sec:edit_result}. 

\begin{table*}[t]
\centering
\resizebox{\linewidth}{!}{
\begin{tabular}{|c|c|c|c|}
\hline
\textbf{Testing type} & \textbf{Input}  & \textbf{Prediction \textbf{w.o.} CREME} & \textbf{Prediction \textbf{w.} CREME} \\
\hline
\textbf{Hasty Answer \uppercase\expandafter{\romannumeral2}}\\
\hline
Paraphrasing & What is the citizenship of the creator of C. Auguste Dupin? &  \textcolor{red}{France} & \textcolor{green}{American} \\
\hline
 Paraphrasing& What is the nationality of the creator of C. Auguste Dupin? & \textcolor{red}{France} &  \textcolor{green}{United States of America} \\
\hline
Paraphrasing & The country where the creator of C. Auguste Dupin is a citizen is &  \textcolor{red}{France} & \textcolor{green}{United States of America} \\
\hline
Generalization & Which city did the creator of C. Auguste Dupin die in? & \textcolor{red}{Paris} & \textcolor{green}{Baltimore, Maryland} \\
\hline
\textbf{Incomplete Reasoning} \\
\hline
Paraphrasing & What is the capital of the country where Sven Väth is a citizen? &  \textcolor{red}{Germany} & \textcolor{green}{Berlin} \\
\hline
 Paraphrasing& In what city is the capital located of the country that Sven Väth is a citizen of? & \textcolor{red}{Germany} &  \textcolor{green}{Berlin} \\
\hline
Generalization & The official language of the country that Sven Väth is a citizen of is &  \textcolor{red}{Germany} & \textcolor{green}{German} \\
\hline
\end{tabular}
}
\caption{
Case study for generalization effect of correcting the (1) \textbf{Hasty Answer \uppercase\expandafter{\romannumeral2}} error: the original input (used for correcting) is ``The country that the creator of C. Auguste Dupin belongs to is”. The original prediction is ``France” (Reference: C. Auguste Dupin is French, while his creator Edgar Allan Poe. is American.); and (2) \textbf{Incomplete Reasoning} error: the original input (used for correcting) is ``The capital of the country that Sven Väth is a citizen of is”. The original prediction is ``Germany” (Reference: Berlin.).
}
\label{tab:showcase1}
\end{table*}

\subsection{Dataset, Baseline and Evaluation Metric}
\paragraph{Dataset}
The dataset $\mathcal{D}_{edit}$ we use for editing and evaluating LLMs is built based on the $\mathcal{D}_{gap}$ filtered in Section~\ref{sec:infer_dataset}.
For each example in $\mathcal{D}_{edit}$, it has the following fields:
(1) \textbf{Original} input $I_o$ is a cloze test form of the compositional two-hop query. Accordingly, we also have the correct answer (ground-truth) and the originally predicted wrong answer for $I_o$: $A_o$ and $\widetilde{A_o}$, respectively\footnote{e.g.,for the fourth case in Table~\ref{tab:error_type}:$A_o$=England; $\widetilde{A_o}$=Italy.}. In the experiment, we use $I_o$ and its corresponding second-hop query to edit the LLM.
(2) \textbf{Paraphrasing} input $I_p$ is a paraphrase of $I_o$. Note that $A_o$ and $\widetilde{A_o}$ are also applicable to $I_p$.
(3) \textbf{Generalization} input $I_g$ is a compositional two-hop query where its first-hop sub-knowledge is shared with $I_o$ while the second-hop sub-knowledge is different from $I_o$. We denote the correct answer for $I_g$ is $A_g$.
(4) \textbf{Irrelevant} input $I_i$ is a compositional two-hop query that is irrelevant to $I_o$ and does not share the final answer with $I_o$.
Detailed information about $\mathcal{D}_{edit}$ is available in Appendix~\ref{appendix:datasets}.
\paragraph{Baseline} 
We choose two related works in the field of correcting compositional reasoning errors through manipulating the inner workings of LLMs: \textbf{Memory Injection}~\cite{memoryinjections_blackboxnlp2023} and \textbf{CoT-PatchScopes}~\cite{ghandeharioun2024patchscopes} as our baselines.
Memory Injection enhances the compositional reasoning through explicitly injecting the implicit reasoning result (so-called ``memory”) into the hidden states in the residual stream. 
CoT-PatchScopes corrects the compositional reasoning through mimicking the 
noted
Chain-of-Thought (CoT) reasoning~\cite{cot_nips2022} to re-route forward computation. 
Besides, we also compare CREME with \textbf{ROME}~\cite{meng2022locating}, a state-of-the-art model editing method.
Detailed implementations are available in Appendix~\ref{appendix:implementations}.
\paragraph{Evaluation Metric} 
In order to comprehensively validate the effectiveness of CREME, we propose four evaluation metrics: \textbf{\textit{Correction}}, \textbf{\textit{Paraphrasing}}, \textbf{\textit{Generalization}} and \textbf{\textit{Specificity}}. Following ~\citet{memoryinjections_blackboxnlp2023}, all the metrics are formulated on the basis of Improvement Percentage (\textbf{IP}), which is calculated as $\text{IP}(I,A)=\frac{p_{\mathcal{M}^*}(A|I)-p_\mathcal{M}(A|I)}{p_\mathcal{M}(A|I)}$.
This formula quantifies the enhancement in prediction probability of an answer $A$ given an 
input query $I$, facilitated by the post-edit LLM $\mathcal{M}^*$ in comparison to the pre-edit LLM $\mathcal{M}$.
Specificially, \textit{Correction} quantifies $\text{IP}(I_o,A_o)$ (larger is better);
\textit{Paraphrasing} is $\text{IP}(I_p,A_o)$ (larger is better);
\textit{Generalization} is $\text{IP}(I_g,A_g)$ (larger is better)
and \textit{Specificity} is $\text{IP}(I_i,A_o)$ (smaller is better). 
CoT-PatchScopes, due to its nature of input-dependent, only fits the \textit{Correction} evaluation.
We report the average results over $\mathcal{D}_{edit}$ in Section~\ref{sec:edit_result}.
Note that to handling the multiple tokens in the answer, we calculate the probabilities for all of the tokens in the predicted probability distribution and report the mean value.
\subsection{Experiment Results}
\label{sec:edit_result}

The main experiment results are shown in Table~\ref{tab:edit_main}. For brevity, we omit $\times 100\%$ for each IP value.
We observe that CREME achieves better performance than baselines on all metrics, not only achieving notable improvement on $I_o$ (the query used for editing), but also effectively generalizing to $I_p$ (paraphrased queries). 
Interestingly, editing with $I_o$ also improves (at most $+366\%$) the compositional reasoning on $I_g$ (only sharing first-hop knowledge with $I_o$), demonstrating the effectiveness of CREME on generating proper implicit reasoning results in middle layers.
Besides, the Specificity score of CREME is low, showing that the CREME does not aimlessly improve the probability of predicting $A_o$ for irrelevant inputs $I_i$.
In comparison, the Correction score of Memory Injection ($+221\%$ for LLaMA-2) is almost the same with the original paper\footnote{Nonetheless, it still falls far behind CREME. Given that both CREME and Memory Injection aim to enhance the information of implicit reasoning results encoded in intermediary hidden states, we attribute the efficacy of CREME to its compatibility with models.} while we find it is less effective to generalize to $I_p$ and $I_g$. Moreover, its high Specificity score implies its shortcoming of aimlessly improving the probability of predicting $A_o$.
We also show $\text{IP}(I_o,\widetilde{A_o})$ in Figure~\ref{fig:wrong}. A good correction method should have little positive improvement on predicting the wrong answer $\widetilde{A_o}$. We observe that $p(\widetilde{A_o}|I_o)$ approximately remains unchanged with CREME, while is apparently enlarged with Memory Injection and PatchScopes.

One natural concern arises regarding the sufficiency of \textbf{\textit{Correction} and \textit{Paraphrasing} metrics in practice}. To this end, we evaluate the probability of an event where the probability of predicting $A_o$ exceeds that of predicting $\widetilde{A_o}$: $p(A_o)\!>\!p(\widetilde{A_o})$. We compare CREME against baselines using this new metric and two types of input ($I_o$ and $I_p$) in Table~\ref{tab:edit_greater}. The results underscore CREME's efficacy in significantly improving the event probability, thereby outperforming the unedited LLM and establishing a considerable lead over the two baselines.

Although CREME is not comparable to traditional model editing methods (the latter require $A_o$ for editing, while CREME does not), we compare CREME with a well-regarded model editing method: ROME~\cite{meng2022locating} for a comprehensive investigation.
The results\footnote{\textit{Correction} and \textit{Paraphrasing} scores are using the event probability of $p(A_o|I)>p(\widetilde{A_o}|I)$.} are shown in Table~\ref{tab:edit_rome}. Our findings reveal that while ROME marginally surpasses CREME in terms of the \textit{Correction} score of ROME -- attributable to ROME's direct application of 
$A_o$ for editing and its optimization procedure designed to entirely fit $p(A_o)$ -- CREME performs obviously better than ROME in paraphrased, generalization and irrelevant cases. This highlights the effectiveness of CREME on correcting compositional reasoning.

To make readers have better sense of the realistic effect of using CREME to improve LLMs' compositional reasoning performance, we also report the (1) decreasement percentage of $\log\text{PPL}$ values\footnote{\url{https://huggingface.co/docs/transformers/en/perplexity}} (of predicting correct answers) and (2) prediction accuracy in paraphrasing and generalization testing cases before and after applying CREME in Table~\ref{fig:edit_final_predict}.

In Figure~\ref{fig:edit_layer}, we show the effects of \textbf{editing different layers}, where results align well with the results of the locating experiment (Figure~\ref{fig:locating_dataset}).
\begin{table}[t]
\centering
\resizebox{0.5\textwidth}{!}{
\begin{tabular}{lcccc}
\hline
\textbf{Evaluation Metrics} & \textbf{C}($\uparrow$)  & \textbf{P}($\uparrow$) & \textbf{G}($\uparrow$) & \textbf{S}($\downarrow$) \\
\hline
\hline
LLaMA-2-7B & $3.2\%$ & $2.3\%$ & $13.1\%$  & $0.3\%$ \\
\hline
\textit{CoT-PatchScopes} & $+1.20$ & --  & --   & --  \\
\textit{Memory Injection} & $+2.21$ & $+0.30$ & $+0.32$  & $+26.72$ \\
\textit{CREME} \footnotesize{(\textbf{Ours})} & $+\mathbf{17.0}$ & $+\mathbf{7.99}$ & $+\mathbf{1.27}$ & $+\mathbf{0.86}$ \\
\hline
\hline
OpenAlpaca-3B & $7.2\%$ & $7.0\%$ & $13.5\%$  & $0.6\%$ \\
\hline
\textit{CoT-PatchScopes} & $+0.91$ & --  & --   & --  \\
\textit{Memory Injection} & $+0.98$ & $+0.45$ & $+0.75$  & $+2.93$ \\
\textit{CREME} \footnotesize{(\textbf{Ours})} & $+\mathbf{43.3}$ & $+\mathbf{23.71}$ & $+\mathbf{3.61}$ & $+\mathbf{1.24}$ \\
\hline 
\end{tabular}
}
\vspace{-0.1in}
\caption{
CREME versus baselines with the proposed four metrics: \textbf{C} for ``Correction”, \textbf{P} for ``Paraphrasing”, \textbf{G} for ``Generalization” and \textbf{S} for ``Specificity”. Note that the values in the table is averaged improvement percentage (i.e., we calculate the improvement percentage for each single case and then do average over the dataset.) and hence it is meaningless to calculate values like $7.2\%\times(1+43.3)=318.96\%>100\%$.
}
\vspace{-0.1in}
\label{tab:edit_main}
\end{table}

\begin{table}[t]
\centering
\resizebox{0.5\textwidth}{!}{
\begin{tabular}{lcc}
\hline
\textbf{Input Types} & Correction Input $I_o$  & Paraphrasing Input $I_p$  \\
\hline
\hline
LLaMA-2-7B\\
\hline
\textit{Original}& $59.5\%$ & $35.7\%$  \\
+\textit{CoT-PatchScopes}& $53.0\%$ & --  \\
+\textit{Memory Injection}& $63.0\%$ & $40.3\%$  \\
+\textit{CREME}\footnotesize{\textbf{(Ours)}}& $\mathbf{87.5}\%$ & $\mathbf{52.9}\%$  \\
\hline
\hline
OpenAlpaca-3B \\
\hline
\textit{Original} & $58.0\%$ & $42.7\%$  \\
+\textit{CoT-PatchScopes}& $57.3\%$ & --  \\
+\textit{Memory Injection}& $58.7\%$ & $43.8\%$  \\
+\textit{CREME}\footnotesize{\textbf{(Ours)}} & $\mathbf{95.3}\%$ & $\mathbf{70.5}\%$ \\
\hline 
\end{tabular}
}
\vspace{-0.1in}
\caption{
The event probability of $p(A_o)>p(\widetilde{A_o})$.
}
\vspace{-0.1in}
\label{tab:edit_greater}
\end{table}



%% file: Working_Draft/sec7_related_work.tex
\section{Related Work}
\paragraph{Compositional Reasoning of LLMs}
LLMs fail to solve a large proportion of compositional multi-hop questions, even successfully solving all their single-hop sub-questions~\cite{measuring_emnlp2023,dziri2023faith}. 
Early works towards mitigating this issue typically prepend crafted demonstration exemplars containing the ``thought process” of solving the compositional query step-by-step and encourage LLMs to imitate the process via in-context learning~\cite{nye2021work,cot_nips2022,zhou2023leasttomost,drozdov2023compositional,measuring_emnlp2023}. 
Recent works turn to inspect the inherent compositional reasoning mechanism~\cite{hou-etal-2023-towards} of LLMs. 
 ~\citet{memoryinjections_blackboxnlp2023} manually injects implicit reasoning results into LLMs at the middle layers to correct compositional reasoning failures. ~\cite{ghandeharioun2024patchscopes} fixes compositional reasoning errors through re-routing inner hidden representations in the computation graph to mimic chain-of-thought reasoning process.
Nonetheless, their interventions in the reasoning process are rough so that the improvement is limited and hardly generalize to other related queries.
To this end, we elaborately analyze the cause of compositional reasoning failures, locate a small set of parameters in LLMs that are responsible for such failures and precisely edit them to correct such failures. 
The work and ~\citet{yang2024large} are concurrent, where both of the two works show empirical evidence that LLMs can latently perform multi-hop reasoning internally. 

%% file: Working_Draft/sec8_conclusion.tex
\section{Conclusion}
In this paper we study and patch the compositional reasoning of LLMs.
Through examining failure instances and conducting diverse analysis experiments, we demonstrate successful compositional reasoning within LLMs hinges on its awareness of generating and leveraging implicit reasoning results.
Moreover, we locate few important MHSA modules in LLMs that are responsible for properly generating and leveraging implicit reasoning results via causal mediation analysis.
To this end, we propose CREME, to compositional reasoning failures via editing the located MHSA parameters and empirically demonstrate its superiority.
\section*{Limitations}
\paragraph{Technique}
Part of our observation and experiments in Section~\ref{sec3:analyzing compositional reasoning} and Section~\ref{sec4:inference and errors} are on the basis of Logit Lens~\cite{LogitLens2020}. Though Logit Lens is a widely used tool for analyzing the inner workings of language models~\cite{geva-etal-2022,geva-etal-2023-dissecting,dar-etal-2023-analyzing,memoryinjections_blackboxnlp2023,katz-belinkov-2023-visit,ram-etal-2023-token}, we acknowledge that it is only an approximate way to interpret the information in the inner hidden states of the LLMs~\cite{belrose2023eliciting}. Nonetheless, the residual stream architecture of Transformers guarantees that Logit Lens makes sense to a large extent. In our experiments, we try to conduct experiments with different techniques for the \textbf{cross-validation} of our observations and conclusions (By way of example, the observations in the locating experiments (Section~\ref{sec5:locating_important_modules}) to some extent validate the observations of the intervention experiments in Section~\ref{sec3:causal intervention}).
\paragraph{LLM}
Due to the constraints of available computation resource, we are able to conduct most of our experiments with LLMs of seven billion scale (LLaMA-2-7B~\cite{touvron2023llama}) and three billion scale (OpenAlpaca-3B~\cite{openalpaca}). Both of these two LLMs are fully open-sourced and popular in academic community and real-world applications~\cite{wu2023interpretability,wang2024how,hou-etal-2023-towards,li2023inferencetime}. In the future work, we aim to validate our conclusions on LLMs of larger scale.
\paragraph{Task}
In this work, we mainly focus on the task of the compositional reasoning on factual knowledge, which is generally pursued by lots of research works~\cite{misra-etal-2023-triggering,measuring_emnlp2023,zhong-etal-2023-mquake,memoryinjections_blackboxnlp2023}. We aim to validate our main conclusion about the significance of implicit reasoning results in the compositional reasoning process in other types of compositional reasoning task~\cite{lu2023chameleon, hou-etal-2023-towards}(e.g., Arithmetic Reasoning for multiple operands) in the future work.
\section*{Ethical Considerations}
We study the inner workings for the compositional reasoning of LLMs, which helps the black-box LLMs become more transparent and trustworthy~\cite{räuker2023transparent}.
The CREME method introduced in this work is originally designed for correcting the compositional reasoning failures of LLMs. CREME only require slightly update a small set of parameters in LLMs and can generalize to a number of related queries (paraphrased queries or compositional queries sharing first-hop knowledge with the query used for conducting CREME).
However, just like traditional model editing methods~\cite{de-cao-etal-2021-editing, mitchell2022fast, meng2022locating, meng2023massediting,bi2024decoding}, it may also be utilized to insert inaccurate (or out-of-date) information into the pretrained LLMs, potentially resulting in negative influences in real-word applications of LLMs such as information retrieval~\cite{lian2020personalized} and recommendation~\cite{lian2014geomf,lian2020lightrec}.

\section*{Acknowledgements}
Special thanks to Dr.Xiting Wang from Renmin University of China for her helpful comments to the early version of the work.
We also sincerely thank the anonymous reviewers for their insightful suggestions to this work. The work was supported by grants from the National Key R\&D Program of China (No.2021ZD0111801), the Research Grants Council of the Hong Kong SAR under Grant GRF 11217823 and Collaborative Research Fund C1042-23GF, the National Natural Science Foundation of China under Grant 62371411 and InnoHK initiative the Government of the HKSAR,Laboratory for AI-Powered Financial Technologies.

%% file: Working_Draft/Appendix.tex
\section{Related Works on Mechanistic Interpretability and Model Editing}
\paragraph{Mechanistic Interpretability and Model Editing}
Mechanistic Interpretability, interpreting inner workings of LLMs, is drawing an increasing attention of NLP researchers. Logit Lens~\cite{LogitLens2020} is proposed to interpret hidden states at the middle layers of LLMs via projecting them to the output vocabulary space with the LM head. Subsequent works~\cite{geva-etal-2021-transformer,geva-etal-2022, dar-etal-2023-analyzing, katz-belinkov-2023-visit} further explain how LLMs build precise next token predictions. Another line of mechanistic interpretability works focus on inspecting factual knowledge encoded in the LLMs: they first locate such factual knowledge in pretrained LLMs~\cite{dai-etal-2022-knowledge,geva-etal-2023-dissecting,li2023inferencetime} and then edit them through updating a small set of parameters of LLMs~\cite{meng2022locating, meng2023massediting,hase2023does}, which is so-called ``locate-then-edit” model editing~\cite{ju2023klob}. 
In this paper, we shed light on the mechanism of compositional reasoning on factual knowledge and borrow the idea from ``locate-then-edit” model editing to locate the root and correct it to patch compositional reasoning failures of LLMs.
\section{Datasets}
\label{appendix:datasets}
\paragraph{Dataset for Non-Editing Experiments}
Here we mainly introduce the dataset we use for Non-Editing experiments (including inspecting experiments in Section~\ref{sec3:inspecting}, intervention experiments in Section~\ref{sec3:causal intervention}, inference experiments in Section~\ref{sec4:inference and errors} and locating experiments in Section~\ref{sec5:locating_important_modules}.)
The dataset $\mathcal{D}$ we use in this paper is sourced from ~\cite{zhong-etal-2023-mquake}, a dataset containing plenty of high-quality compositional multi-hop reasoning cases. For the ease of our study and following the setting of ~\cite{measuring_emnlp2023}, we collect 1,000 two-hop knowledge items (each with its two single sub-knowledge) as the base of our dataset. For each datum in the dataset, it contains the following component: 
(1) four paraphrased compositional two-hop knowledge $(s_1,r_1,o_1)\oplus (s_2,r_2,o_2) (o_1 = s_2)$ queries: one of them is in Cloze-Test form and the other three is in Question form;
(2) two paraphrased first-hop sub-knowledge $(s_1,r_1,o_1)$ queries: one is in Cloze-Test form and another is in Question form;
(3) two paraphrased second-hop sub-knowledge $(s_2,r_2,o_2)$ queries: one is in Cloze-Test form and another is in Question form;
and (4) the results for compositional reasoning: the intermediate \textbf{implicit reasoning result} $o_1$ (meanwhile is the answer for the first-hop queries) and the final \textbf{explicit reasoning result} $o_2$ (meanwhile is the answer for the second-hop queries).
Following ~\cite{meng2022locating, geva-etal-2023-dissecting, zhong-etal-2023-mquake, measuring_emnlp2023}, We use the Question form queries in the inference experiment (Section~\ref{sec4:inference and errors}) and Cloze-Test form queries in most of the rest experiments in this paper. 
Below is an example for the datum in our dataset.
\begin{lstlisting}
{
    "compositional question query": [
            "Which writer's country of citizenship is the same as the author of \"Misery\"?",
            "What country does the author of \"Misery\" and another writer share their citizenship?",
            "What is the nationality of the author of \"Misery\"?"
    ],
    "compositional cloze query": "The nationality of the author of \"Misery\" is",
    "first-hop question query": "Who is the author of \"Misery\"?"
    "first-hop cloze query": "The author of \"Misery\" is"
    "second-hop question query": "What is the nationality of Stephen King?"
    "second-hop cloze query": "The nationality of Stephen King is",
    "compositional answer": "United States of America", // explicit reasoning result
    "first-hop answer": "Stephen King", // implicit reasoning result
    "second-hop answer": "United States of America"
}
\end{lstlisting}
\paragraph{Dataset for Editing Experiments}
Here we mainly introduce the dataset we use for conducting and evaluating CREME (Correcting Compositional Reasoning via Model Editing) in Section~\ref{sec:creme}.
The dataset $\mathcal{D}_{edit}$ we use for editing and evaluating LLMs is built on top of the dataset $\mathcal{D}_{gap}$ filtered in Section~\ref{sec:infer_dataset}: for a LLM $\mathcal{M}$: we focus on the example that $\mathcal{M}$ succeeds to predict the correct answer given any of single-hop inputs in it while fails to correctly predict the answer for the corresponding compositional two-hop input in it. In this section, we are going to correct these compositional reasoning failures. Specifically, for each example in $\mathcal{D}_{edit}$, it has the following components:
(1) \textbf{Original} input $I_o$, refers to a cloze test form of the compositional two-hop knowledge mentioned above. Accordingly, we also have the correct answer (ground-truth) and the originally predicted wrong answer for $I_o$: $A_o$ and $\widetilde{A_o}$, respectively\footnote{E.g., for the first case in Table~\ref{tab:error_type}: $A_o$=England; $\widetilde{A_o}$=Italy.}.
(2) \textbf{Paraphrasing} input $I_p$, refers to a paraphrase (e.g., cloze test $\rightarrow$ question) of $I_o$ (we collect 3.39 $I_p$ for each $I_o$ in average). Note that $I_p$ shares the $A_o$ and $\widetilde{A_o}$ with $I_o$.
(3) \textbf{Generalization} input $I_g$, refers to a verbalized compositional two-hop knowledge where the first-hop sub-knowledge is shared with $I_o$ and the second-hop sub-knowledge is different from $I_o$ (we collect 2.64 $I_g$ for each $I_o$ in average). We denote the correct answer for $I_g$ is $A_g$.
(4) \textbf{Irrelevant} input $I_i$, refers to a verbalized compositional two-hop knowledge that is irrelevant to $I_o$ and does not share the final answer with $I_o$ (we collect 9.49 $I_i$ for each $I_o$ in average).
Below is an example for the dataset (corresponding to the \textbf{Incomplete Reasoning} type of errors in Section~\ref{sec4:identify error types}).
\begin{lstlisting}
{
    "Original Input": "The capital of the country that Lou Pearlman is a citizen of is",
    "Correct Answer for I_o": "Washington, D.C.",
    "Predicted Wrong Answer for I_o": "United States of America",
    "Paraphrasing Input":[
        "What is the capital of the country to which Lou Pearlman belonged?",
        "Which city serves as the capital of the country where Lou Pearlman was a citizen?",
        "In which city is the capital of the country where Lou Pearlman had citizenship?",
        "The capital of the country to which Lou Pearlman belonged is",
        ...
    ],
    "Generalization Input": [
        "The official language of the country that Lou Pearlman is a citizen of is",
        "What is the official language of the country that Lou Pearlman is a citizen of?",
        ...
    ],
    "Generalization Answer": [
        "American English",
        "American English",
        ...
    ]
    "Irrelevant Input": [
        "Which continent is the country that Emma Bunton is a citizen of located in?",
        "The official language of the country that Thierry Mugler is a citizen of is",
        ...
    ],
    "Irrelevant Answer": [
        "Europe",
        "French",
        ...
    ]
}
\end{lstlisting}
\section{Language Models}
\label{appendix:LLMs}
\subsection{LLM Architecture}
\label{appendix:llm_arch}
Current Large Language Models (LLMs, in this paper, we conduct most of the experiments with two popular and open-sourced LLMs\footnote{Due to the page limit, we sometimes present the results with one of them while readers can find the rest results in Appendix~\ref{appendix:additional results}.}: LLaMA-2-7B~\cite{touvron2023llama} and OpenAlpaca-3B~\cite{openalpaca, alpaca}.) are mostly built on the basis of traditional Transformer~\cite{transformer_nips2017} (Decoder). They are typically consist of an embedding layer $E$, an output language model (LM) head $W_u$ and a stack of repetitive Transformer blocks between $E$ and $W_u$.
\paragraph{Embedding Layer} 
Given a tokenized input $inp=[t^1, t^2, ..., t^N]$, where each $t_i\ (1\le i\le N)$ is a one-hot vector of $|V|$ ($V$ is the vocabulary set) dimensions, the embedding layer is actually an embedding matrix $E\in \mathbb{R}^{|V| \times d}$, projecting the input sparse one-hot vectors into $d$-dimensional hidden space: $inp\cdot E = [h_0^1, h_0^2, ..., h_0^N]$.  $h_0^i\ (1\le i\le N)\in \mathbb{R}^{d}$ is the initial hidden state that is forwarded into the first Transformer block (Note that we omit the description for the rotary positional embedding (RoPE)~\cite{su2023roformer} added at each Transformer block of the network). 
\paragraph{Transformer Block}
A Transformer block (or a Transformer layer) typically has two sub-modules: a Multi-Head Self-Attention (MHSA) layer and a Multi-Layer Perceptron (MLP) layer.
We denote the hidden states at the input and output of the $l$-th ($1\le l \le L$) Transformer Block are $h_{l-1}$ and $h_l$ respectively (Since hidden states of all token positions are forwarded parallelly, we define $h_l \triangleq [h_l^1, h_l^2, ..., h_l^N] \in \mathbb{R}^{N\times d}$ to represent the whole hidden states of the $l$-th layer.). Then we have:
\begin{equation}
    h_l = h_{l-1} + a_{l} + m_{l} \in \mathbb{R}^{N\times d}
\end{equation}
where $a_{l}$ and $m_{l}$ refer to the MHSA output and the MLP output.

MHSA layer of $l$-th Transformer block contains four matrices: $W_Q^l,W_K^l,W_V^l,W_O^l\in\mathbb{R}^{d\times d}$. Let $H$ denote the number of attention heads. Then the parameters in each matrix can be equally divided into $H$ parts: each of them is an individual attention head (e.g., for the $j$-th head, $1\le j \le H$): $W_Q^{l,j},W_K^{l,j},W_V^{l,j} \in \mathbb{R}^{d\times \frac{d}{H}}$ and $W_O^{l,j}\in \mathbb{R}^{\frac{d}{H} \times d}$. Then we first compute the attention value for the $j$-th head: ($M\in\{0,1\}^{N\times N}$ is the attention mask matrix)
\begin{align*}
    & A^{l,j} = \text{softmax}(\frac{(h_{l-1}W_Q^{l,j})(h_{l-1}W_K^{l,j})^T}{\sqrt{d/H}}\odot M)\\
    & \text{head}_{l}^j = A^{l,j}(h_{l-1}W_V^{l,j}) \in \mathbb{R}^{N\times d/H} \\
\end{align*}
The final output of the MHSA $a_{l}$ is to concatenate these heads together:
\begin{align}
    a_{l} = \text{Concat}(\text{head}_{l}^1, \text{head}_{l}^2, ..., \text{head}_{l}^H) W_O^l \in \mathbb{R}^{N\times d}
    \label{eq:attention_output}
\end{align}
MLP layer of $l$-th Transformer block contains two matrices: $W_{up}\in\mathbb{R}^{d\times d'}$, $W_{down}\in\mathbb{R}^{d'\times d}$ (in LLaMA-2~\cite{touvron2023llama}, $d'=\frac{8}{3}d$) and a non-linear activation function SwiGLU~\cite{shazeer2020glu} $\sigma$. The output of the MLP $m_l$ can be computed as follows:
\begin{align*}
    m_{l}=\sigma((a_{l}+h_{l-1})W_{up})W_{down} \in \mathbb{R}^{N\times d}
\end{align*}
\paragraph{LM Head}
Let us denote the output of the last Transformer block (at the position of last token) is $h^N_L$ (for LLaMA-2-7B: $L=32$; for OpenAlpaca-3B: $L=26$.). The LM head is a matrix $W_u\in\mathbb{R}^{d\times |V|}$ to project the hidden state $h^N_L\in \mathbb{R}^{d}$ back to the output vocabulary space (probability distribution over the vocabulary set $V$) to predict the next token: 
\begin{align}
    p^N_L = \text{softmax}(h^N_L W_u)
\end{align}
\subsection{LLaMA-2}
LLaMA-2~\cite{touvron2023llama} is a collection of pretrained and fine-tuned generative text models ranging in scale from 7 billion to 70 billion parameters. In this paper, due to the computation resource restraints, we focus on the 7 billion version: LLaMA-2-7b-hf\footnote{\url{https://huggingface.co/meta-llama/Llama-2-7b-hf}}, which is a popular open-sourced LLM in both academic researches and industrial applications. LLaMA-2-7B has 32 layers (32 transformer blocks), a vocabulary size of 32,000 and a hidden dimension of 4,096. In the inference experiments of this paper, we adopt the default generation configuration for LLaMA-2-7B provided by Meta:
\begin{lstlisting}
\\LLaMA-2-7B generation configuration
GEN_CONFIGS["llama2-7b"]={
  "bos_token_id": 1,
  "do_sample": True,
  "eos_token_id": 2,
  "pad_token_id": 0,
  "temperature": 0.6,
  "max_length": 50,
  "top_p": 0.9,
  "transformers_version": "4.31.0.dev0"
}
\end{lstlisting}
\subsection{OpenAlpaca}
OpenAlpaca~\cite{openalpaca} is also an popular instruction-following LLM\footnote{\url{https://github.com/yxuansu/OpenAlpaca}} (fully open-sourced version of Alpaca~\cite{alpaca}). We adopt the 3 billion version: OpenAlpaca-3B\footnote{\url{https://huggingface.co/openllmplayground/openalpaca_3b_600bt_preview}}, for we want to introduce some variation of parameter scales into our experiments. OpenAlpaca-3B has 26 layers (26 transformer blocks), a vocabulary size of 32,000 and a hidden dimension of 4,096. In the inference experiments of this paper, we adopt the default generation configuration for OpenAlpaca-3B provided by ~\citet{openalpaca}:
\begin{lstlisting}
\\OpenAlpaca generation configuration
GEN_CONFIGS["openalpaca-3b"]={
    "do_sample": True, 
    "top_k": 50,
    "top_p": 0.9,
    "generate_len": 128
  "transformers_version": "4.31.0.dev0"
}
\end{lstlisting}
\section{Implementation Details}
\label{appendix:implementations}
\subsection{Intervention}
\label{appendix:implment_causal_intervention}
In the Intervention experiments (Section~\ref{sec3:causal intervention}), a natural worry about the preciseness of the ``intervention” manipulation is whether our intervention will direct affect the probability (observing via Logit Lens) of explicit reasoning results or not. Hopefully, the intervention only works on the ``implicit reasoning result” ($R_i$) while due to the restriction of softmax function, the explicit reasoning result might also be affected by the intervention. In practical, this effect (caused by softmax function) on the ``explicit reasoning result” ($R_e$) is rather insignificant ($\sim 3e-5$) and always increasing the probability (given that the summation of all probabilities over the vocabulary is one, our intervention decrease the probability of $R_i$, naturally improving probabilities for all other tokens.), and hence we do not need to worry about this ``side effect”.
Another potential ``side effect” brought by the intervention is caused for the approximation when solving the inverse matrix with PyTorch \footnote{\url{https://pytorch.org/docs/stable/generated/torch.linalg.inv.html}}. Sometimes, the numerical error brought by the approximation can slightly decrease the probability of $R_e$ (observing via Logit Lens at the intervened layer). To mitigate the possibility that the final effect (in Figure~\ref{fig:debias_dataset}) is attributed to this ``side effect”, We additionally apply the following re-checking procedure (Our aim is that (1): we re-check whether the intervention decrease the probability of $R_e$, and (2): if so, we manually remedy this ``side effect”.).
We first calculate the intervened hidden state $h_l^{*}$:
\begin{equation}
\begin{aligned}
   h_l^{*} & = h_l + (h_l^{*}-h_l) \\
   & = h_l + (v_l^{*}-v_l) W_u^T (W_u W_u^T)^{-1}
\end{aligned}
\end{equation}
Concentrating on $R_e$, we project $h_l^{*}$ to the raw logits $h_l^*W_u$ and check if there is decreasement on the probability of $R_e$.
\begin{equation}
\Delta v_l[j]=\left\{
\begin{aligned}
& v_[j]-(h_l^*W_u)[j], & j\in R_e \\
& 0, & j \in [0..M)/R_e \\
\end{aligned}
\right.
\end{equation}
Then we re-update the hidden state:
\begin{equation}
   h_l^{*,\text{recheck}} = h_l^{*} + \Delta v_l W_u^T (W_u W_u^T)^{-1}
\end{equation}
In Figure~\ref{fig:debias_dataset}, the values is derived from the following procedures: (1) to make positive effect easier to observe, do clipping on the negative effects ($v<0 \rightarrow v=0$) for each single datum; (2) do re-scaling to the effects for each single datum: for effects of different layers, min-max scale them to $[0,1]$; and (3) average the effects over the whole dataset.
\subsection{Inference Experiment}
\label{appendix:inference_prompt}
In line with ~\citet{zhong-etal-2023-mquake} and ~\citet{measuring_emnlp2023}, we adopt the question-form queries to check if LLMs have the single-hop knowledges and whether they can compose them together to answer compositional two-hop questions. The main reason behind using question-form queries is that it is convenient for us to use prompting and In-Context examples to make LLMs directly output the answer. As for the prompt for the question queries, following ~\citet{zhong-etal-2023-mquake}, we prepend eight different demonstrations (namely exemplars) to guide LLMs. Note that, in our experiments, we eliminate the possibility that LLMs directly ``copy” the correct answer from the in-context demonstrations by manually filtering out those demonstrations with the same answer with the questions we want to query. Below is an example for our prompting:
\begin{lstlisting}
Q: In which country was Tohar Butbul granted citizenship? A: Israel\n // eight demonstrations
Q: Who was Nissan 200SX created by? A: Nissan\n
Q: What continent is the country where Prickly Pear grows located in? A: Europe\n
Q: In which country is the company that created Nissan 200SX located? A: Japan\n
Q: Which continent is the country where the director of My House Husband: Ikaw Na! was educated located in? A: Asia\n
Q: What country was the location of the Battle of Pressburg? A: Hungary\n
Q: What is the country of citizenship of Charles II of Spain? A: Spain\n
Q: Who was Chevrolet Biscayne created by? A: Chevrolet\n
Q: What is the name of the head of state of the country that Ellie Kemper is a citizen of? //our query (e.g., compositional question)
\end{lstlisting}
\subsection{Important Module Locating}
We implement our locating method on the basis of Causal Tracing~\cite{meng2022locating}. Following ~\citet{meng2022locating}'s implementation, we also use a ``window” intervention(a few layers before and after the intervened layer). In their original codebase, they set window size to be 10. In our experiments: we find that setting window size to be 2 is enough for us to effectively locate important modules. The locating results shown in Figure~\ref{fig:locating_dataset} adopted the window size of 6.
In Figure~\ref{fig:locating_dataset}, the values is derived from the following procedures: (1) do re-scaling to the indirect effects for each single datum: map the original value (before replacement) to $0$ and map the positive indirect effects to $[0,1]$ and negative indirect effects to $[-1,0]$ in a linear manner; and (2) average the indirect effects over the whole dataset.
\subsection{Model Editing:CREME}
We implement our CREME on the basis of ~\cite{fastedit}. The method is described in Section~\ref{sec:creme}.

The edit objective is the MHSA output matrix of $l$-th layer $W_O^l$. $W_O^l\in\mathbb{R}^{d\times d}$ operates as a key-value store for a set of vector keys $K=[k_1|k_2|...]$ and corresponding vector values $V=[v_1|v_2|...]$, by solving $(W_O^l)^TK=V$.
For a given compositional two-hop query and its corresponding second-hop query, we run the LLM twice: once with the compositional query and once with the second-hop query.
We denote that: in the first pass with compositional query, the \textbf{input} of $W_O^l$ at the last token position is $k_*\in\mathbb{R}^{d\times1}$; in the second pass with the corresponding second-hop query, the \textbf{output} of $W_O^l$ at the last token position is $v_*\in\mathbb{R}^{d\times1}$.
In practice, when calculating $k_*$ and $v_*$, we prepend tens of random tokens to the compositional query and the corresponding second-hop query to mimic context environments,
and get multiple input vectors and output vectors. Then we average input vectors and output vectors of different context environment to get $k_*$ and $v_*$, respectively.
The edited matrix is: $\hat{W_O^l}=W_O^l+(C^{-1}k_*)^T\Lambda^T$ where $C=KK^T$ is a constant to estimate the uncentered covariance of $k$ (with a sample Wikipedia of text) and $\Lambda=(v_*-(W_O^l)^Tk_*)/(C^{-1}k_*)^Tk_*$. 
In our experiments, we also apply a “window” editing (edit a few consecutive layers). In the main experiments, we adopt the window size of 6 in accordance with the locating experiments.
\subsection{Memory Injection}
\label{appendix:memory_injection}
We manually inject memories of implicit reasoning results in to the residual stream of middle layers. 
Note that in the original implementation~\cite{memoryinjections_blackboxnlp2023}, they set a hyper-parameter, magnitude, to control the strength of injection.
In our experiments, we sweep over the possibilities of injecting memories into any single middle layer. For each layer, we search the magnitude from 1 to 10.
As for the matrix used for projecting the implicit reasoning results from the vocabulary space back into the hidden space, we try three different approaches: $W_u^T$ (in line with the original paper), $W_u^{+}$ (Moore–Penrose inverse) and $W^T(WW^T)^{-1}$. We find that $W_u^T$ is always more effective.
\subsection{CoT-PathScopes}
\label{appendix:patchscopes}
We follow the original implementation in Appendix.E. of the PatchScopes paper~\cite{ghandeharioun2024patchscopes}. Rerouting the hidden states (at the last token position) from source layers to target layers. We use the $\frac{p_\mathcal{M}^*(A_o|I_o)-p_\mathcal{M}(A_o|I_o)}{p_\mathcal{M}(A_o|I_o)}$ to select the best source layer and target layer.
\subsection{ROME}
\label{appendix:rome}
We adopt the ~\citet{fastedit}'s implementation of ROME~\cite{meng2022locating}.
The hyperparameters is in line with their original implementation~\cite{fastedit,zhang2024comprehensive}:
\begin{lstlisting}
    layers=[5],
    fact_token="subject_last",
    v_num_grad_steps=20,
    v_lr=1e-1,
    v_weight_decay=1e-3,
    clamp_norm_factor=4,
    kl_factor=0.0625,
\end{lstlisting}
Besides, following the convention of model editing works~\cite{meng2022locating, meng2023massediting}, we also use the Cloze-Test form queries to edit LLMs.
Note that in compositional queries, the ``subject” is usually expressed as the description text containing $s_1$ and $r_1$. We treat the description text as the ``subject” (e.g., ``The sport associated with Giorgio Chinaglia” (association football)).
\section{Additional Results}
\label{appendix:additional results}
\begin{figure}[ht]
    \raggedright
    \includegraphics[width=1.0\linewidth, keepaspectratio=true]{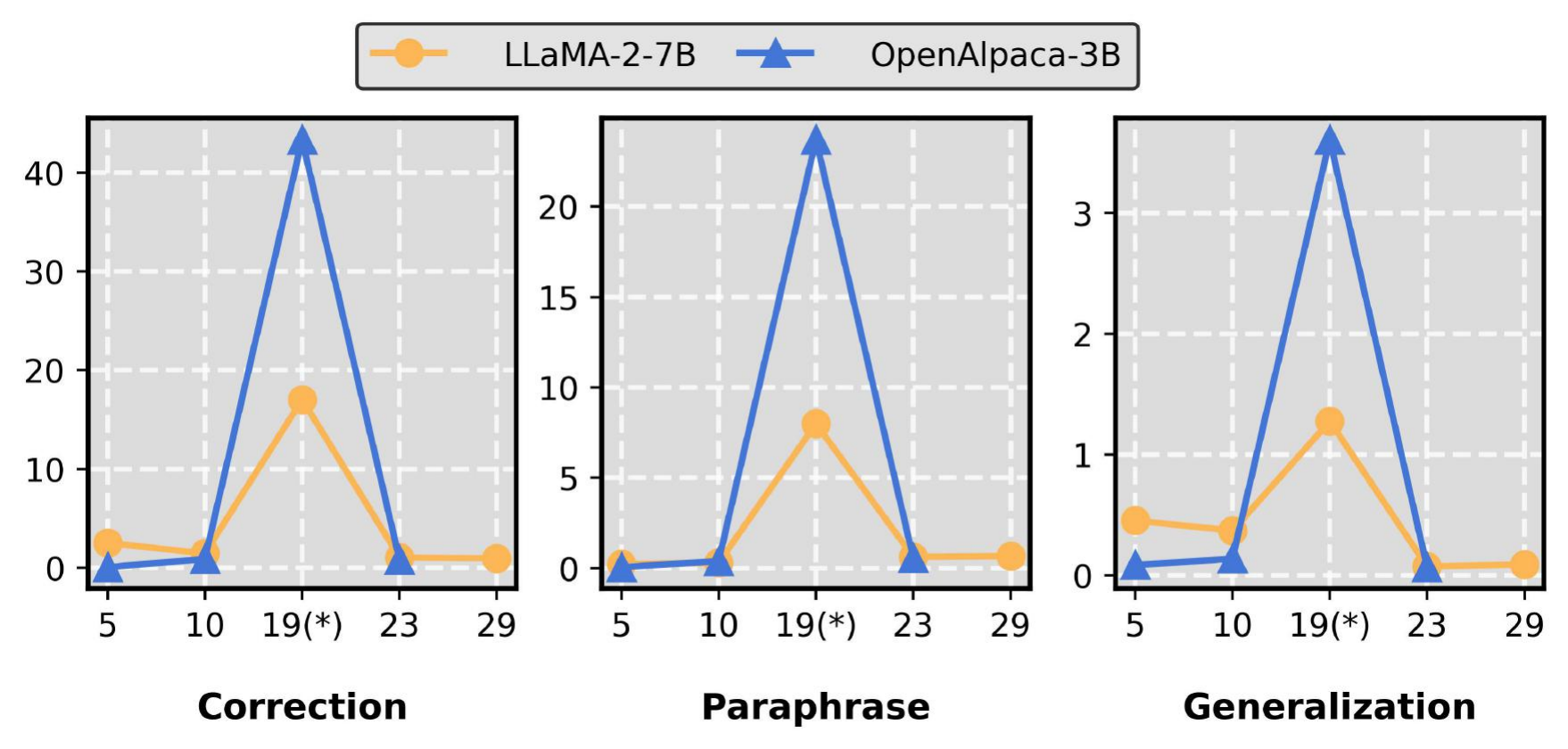}
    \caption{Effects of different editing layers.}
    \label{fig:edit_layer}
\end{figure}

\begin{figure}[ht]
    \raggedright
    \includegraphics[width=1.0\linewidth, keepaspectratio=true]{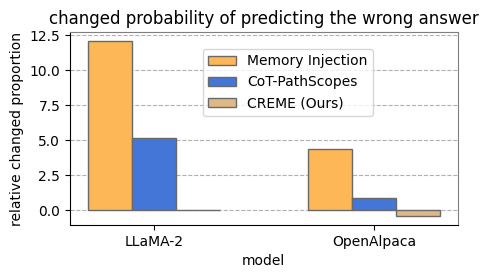}
    \caption{Edit effect on the wrong answer $\widetilde{A_o}$. We anticipate an ideal editing method has little positive effect on predicting $\widetilde{A_o}$.}
    \label{fig:wrong}
\end{figure}
\subsection{Logit Lens Inspecting Results}
\begin{figure}
    \centering
    \subfigure[]{\includegraphics[width=0.48\linewidth]{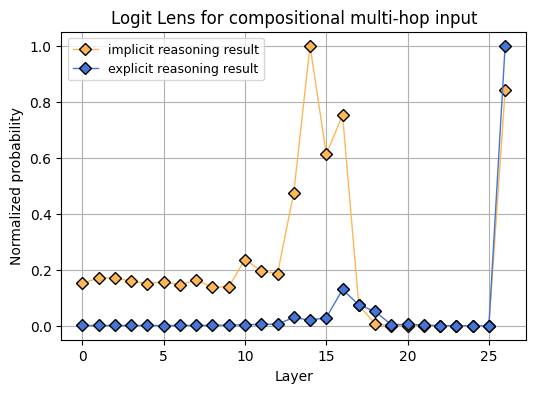}} 
    \subfigure[]{\includegraphics[width=0.48\linewidth]{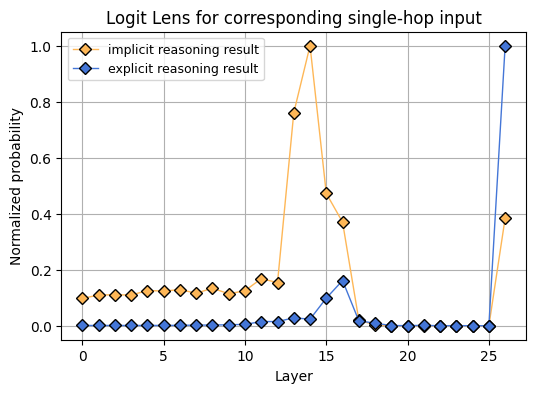}} 
    \caption{Logit Lens inspecting results with OpenAlpaca-3B for a single case. (a) refers to the averaged result for inputs of compositional two-hop knowledge and (b) refers to the averaged result for the inputs of second single-hop knowledge. x-axis refers to the layer; y-axis refers to the 0-1 normalized probability. Yellow line and blue line refers to implicit results and explicit results respectively.}
    \label{fig:logit_lens_alpaca_case}
\end{figure}

\begin{figure}
    \centering
    \subfigure[]{\includegraphics[width=0.48\linewidth]{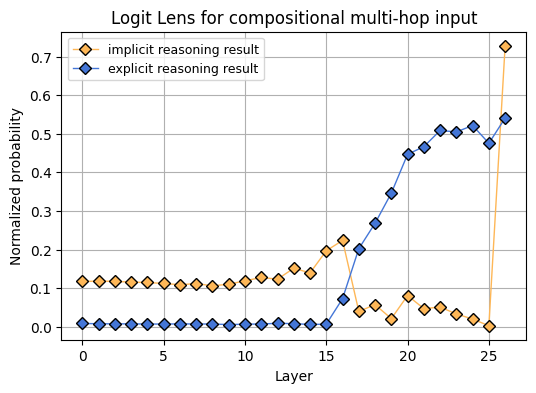}} 
    \subfigure[]{\includegraphics[width=0.48\linewidth]{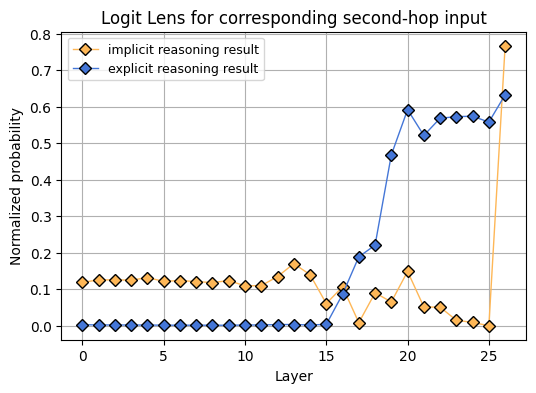}} 
    \caption{Statistical Logit Lens inspecting results with OpenAlpaca-3B. (a) refers to the averaged result for inputs of compositional two-hop knowledge and (b) refers to the averaged result for the inputs of second single-hop knowledge. x-axis refers to the layer; y-axis refers to the 0-1 normalized probability. Yellow line and blue line refers to implicit results and explicit results respectively. The layers of emerging peaks for OpenAlpaca-3B are dispersive in the middle layers.}
    \label{fig:logit_lens_alpaca}
\end{figure}

\begin{figure}
    \centering
    \subfigure[]{\includegraphics[width=0.48\linewidth]{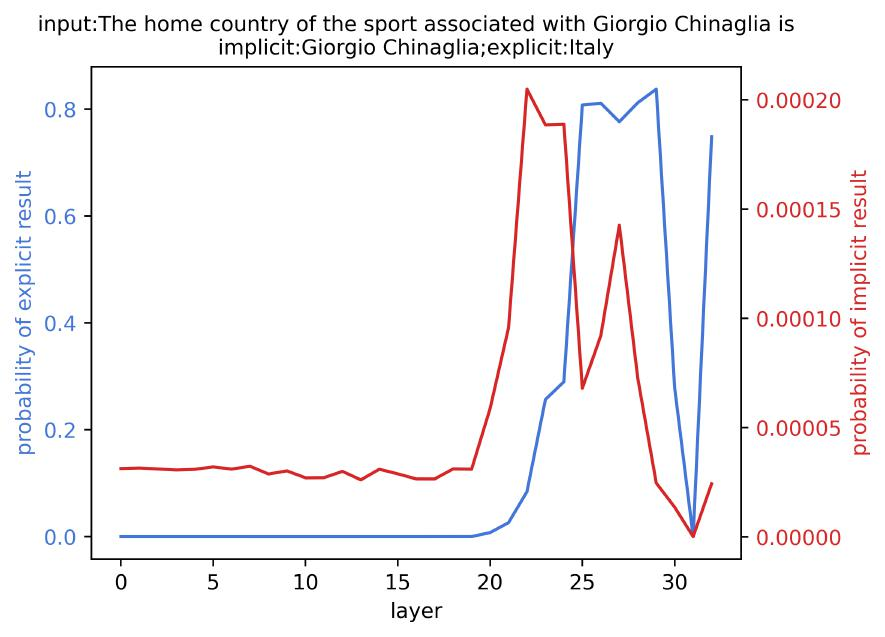}} 
    \subfigure[]{\includegraphics[width=0.48\linewidth]{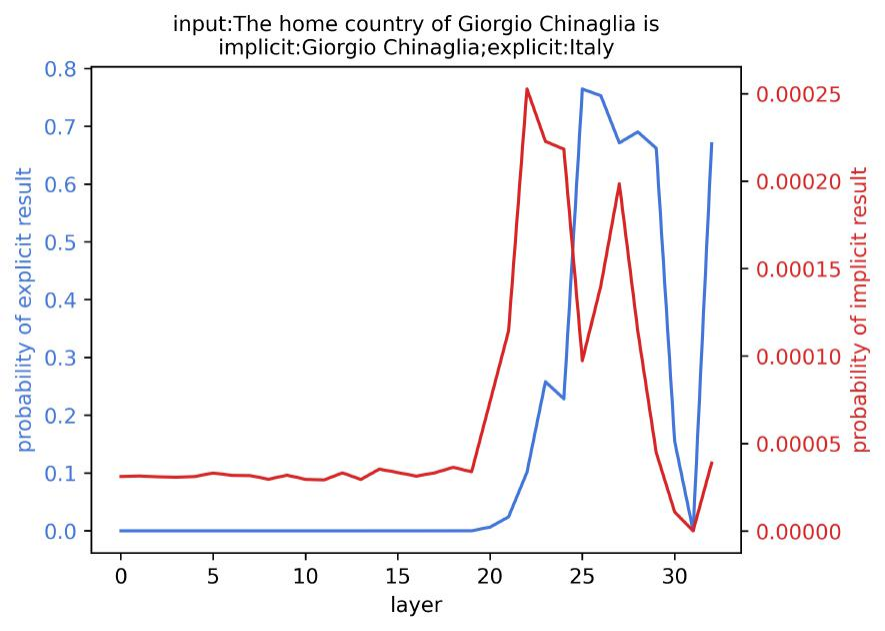}} 
    \caption{Logit Lens results for the \textbf{Hasty Answer \uppercase\expandafter{\romannumeral2}} error type. We investigate the probability of ``Giogrio Chinaglia” (as the implicit reasoning result) and ``Italy” (predicted final answer): the compositional input and the corresponding second-hop input fit well now, implying that the model short-cut ``Giogrio Chinaglia” and ``the home country of” to reason the wrong answer ``Italy”.}
    \label{fig:error_logit_lens_short_cut}
\end{figure}

\begin{figure}
    \centering
    \subfigure[]{\includegraphics[width=0.48\linewidth]{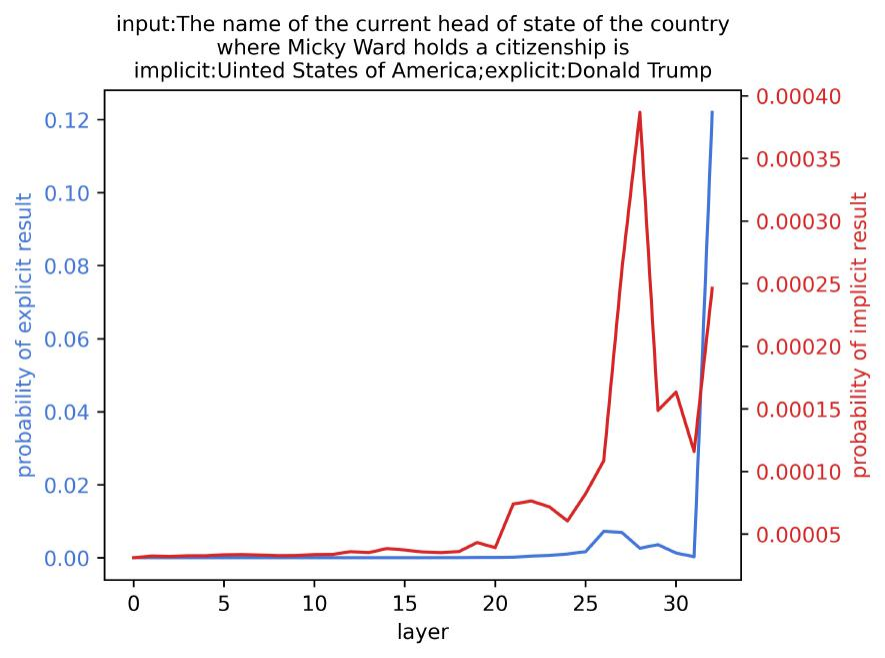}} 
    \subfigure[]{\includegraphics[width=0.48\linewidth]{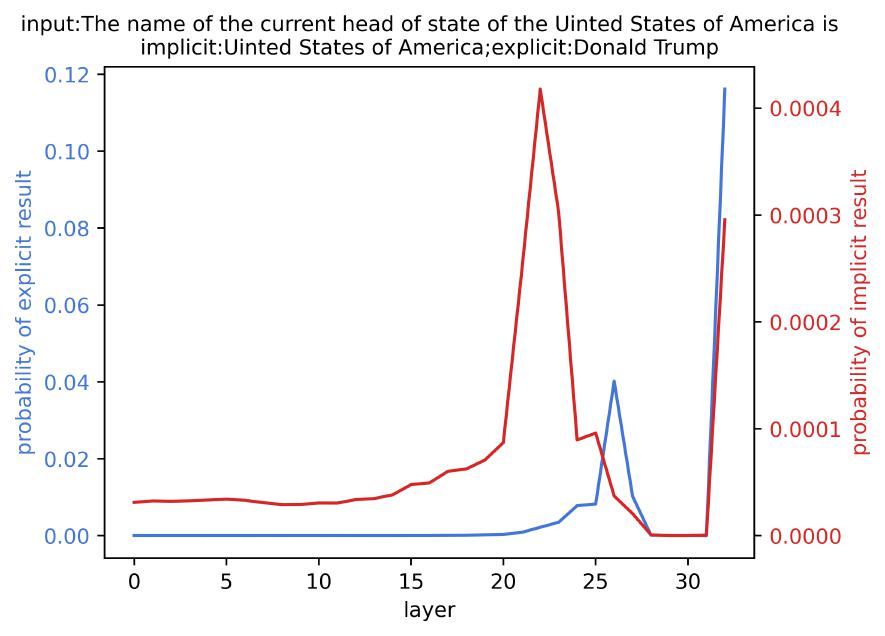}} 
    \caption{A success example in comparison with ``Incomplete Reasoning” error cases. (a) is the inspecting result for compositional two-hop query and (b) is the inspecting result for the reference (corresponding second-hop query). These two results align well (in (a), the implicit reasoning result is properly generated.) and hence the final explicit reasoning results are successfully predicted.}
    \label{fig:error_logit_lens_incomplete_reasoning}
\end{figure}

In this section, we mainly present (1): the statistical Logit Lens inspecting results, (2): a case validating our \textbf{Hasty Answer \uppercase\expandafter{\romannumeral2}} observation (in Section~\ref{sec4:explanation to errors}) and (3): a case validating our \textbf{Incomplete Reasoning} observation (in Section~\ref{sec4:explanation to errors}).
The statistical inspecting result for OpenAlpaca-3B is depicted in Figure~\ref{fig:logit_lens_alpaca}. 
Note that the emerging of ``implicit result” seems not as notable as the results of LLaMA-2-7B in Figure~\ref{fig:logit_lens_init_observation}, the reason is that the layers of emerging peaks for OpenAlpaca-3B are dispersive in the middle layers. We also provide the Logit Lens inspecting results for a single case in Figure~\ref{fig:logit_lens_alpaca_case} for readers' reference.
The cases validating \textbf{Hasty Answer \uppercase\expandafter{\romannumeral2}} and \textbf{Incomplete Reasoning} are depicted in Figure~\ref{fig:error_logit_lens_short_cut} and Figure~\ref{fig:error_logit_lens_incomplete_reasoning}, respectively.
\subsection{Intervention Results}
We present the results for the Intervention experiment (in Section~\ref{sec3:causal intervention}) in Figure~\ref{fig:debias_dataset}.
For each experiment group, we set a \textbf{comparison group} where we intervene on $|R_i|$ tokens that are \textbf{randomly sampled} from $V$. Comparing experiment groups and comparison groups, we observe there exist apparent positive effects ($\mathcal{E}_l>0$) when intervening middle layers (for both LLaMA-2 and OpenAlpaca, positive effects appear in $15\text{-th}\sim20\text{-th}$ layers) for experiment groups, suggesting that the information about $o_1$ may be generated and utilized for generating $o_2$ in these layers.
Meanwhile, there is nearly no notable positive effect for comparison groups across all layers. 
The results verify our hypothesis that the information around implicit reasoning results in middle layers play a role in predicting explicit reasoning results.
\subsection{Memory Injection}
\begin{figure}
    \centering
    \subfigure[]{\includegraphics[width=0.98\linewidth]{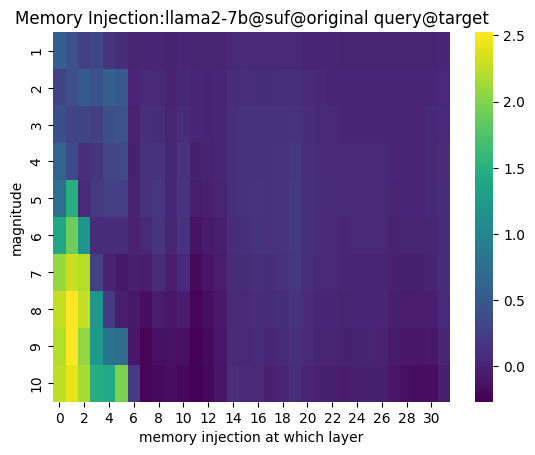}} 
    \subfigure[]{\includegraphics[width=0.98\linewidth]{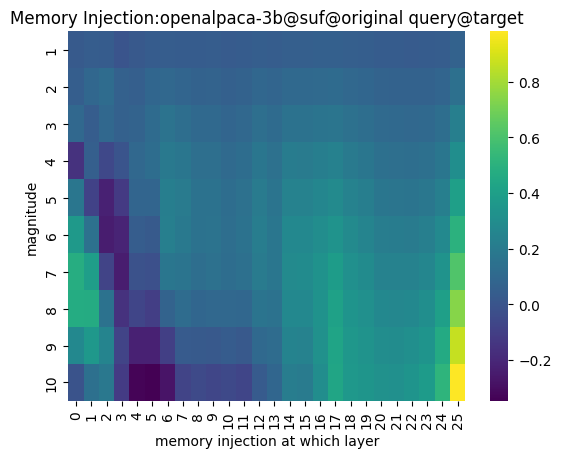}} 
    \caption{(a) depicts the results for LLaMA-2-7B and (b) depicts the results for OpenAlpaca-3B.
    In each subfigure, x-axis refers to the layer of injecting implicit reasoning memories; y-axis refers to the magnitude of injecting memories.
    }
    \label{fig:memory_injection}
\end{figure}

The heatmap of averaged results for Memory Injection~\cite{memoryinjections_blackboxnlp2023} are depicted in Figure~\ref{fig:memory_injection}.
According to this heatmap, for LLaMA-2-7B: we adopt the magnitude of 7 and inject layer of 3, for OpenAlpaca-3B: we adopt the magnitude of 10 and inject layer of 26.
\subsection{PatchScopes}
\begin{figure*}
    \centering
    \subfigure[]{\includegraphics[width=0.48\linewidth]{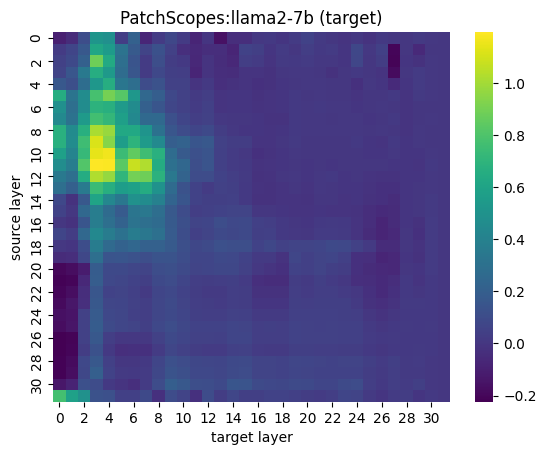}} 
    \subfigure[]{\includegraphics[width=0.48\linewidth]{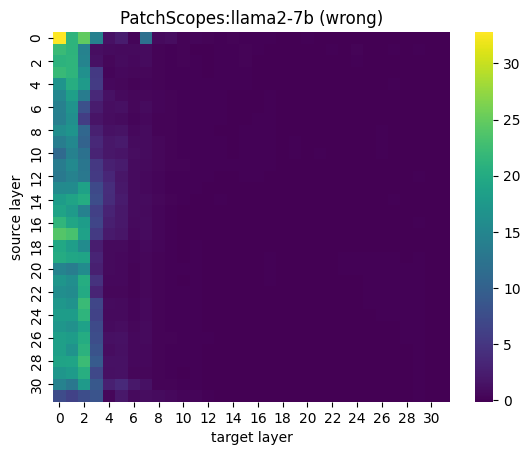}} 
    \subfigure[]{\includegraphics[width=0.48\linewidth]{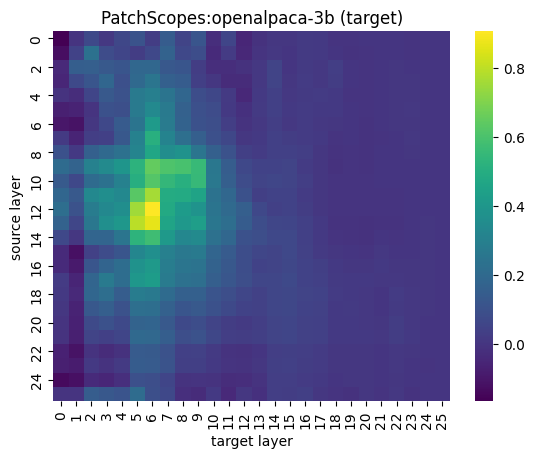}}
    \subfigure[]{\includegraphics[width=0.48\linewidth]{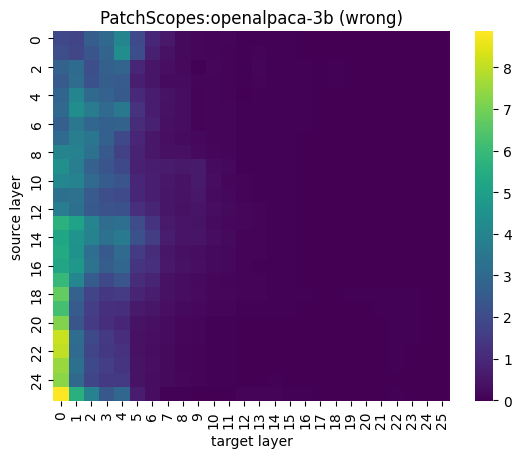}}
    \caption{PatchScopes Results: (a) LLaMA-2-7B, $\frac{p_\mathcal{M}^*(A_o|I_o)-p_\mathcal{M}(A_o|I_o)}{p_\mathcal{M}(A_o|I_o)}$; (b) LLaMA-2-7B, $\frac{p_\mathcal{M}^*(\widetilde{A_o}|I_o)-p_\mathcal{M}(\widetilde{A_o}|I_o)}{p_\mathcal{M}(\widetilde{A_o}|I_o)}$; (c) OpenAlpaca-3B, $\frac{p_\mathcal{M}^*(A_o|I_o)-p_\mathcal{M}(A_o|I_o)}{p_\mathcal{M}(A_o|I_o)}$; (d) OpenAlpaca-7B, $\frac{p_\mathcal{M}^*(\widetilde{A_o}|I_o)-p_\mathcal{M}(\widetilde{A_o}|I_o)}{p_\mathcal{M}(\widetilde{A_o}|I_o)}$.}
    \label{fig:pathscope}
\end{figure*}

The heatmap for PatchScopes~\cite{ghandeharioun2024patchscopes} are depicted in Figure~\ref{fig:pathscope}. The qualitative are basically in align with the original paper: positive effects distributed in the area where the source layer is larger than the target layer. According to this heatmap, for LLaMA-2-7B: we set the source layer to be 12 and the target layer to be 4, for OpenAlpaca-3B: we set the source layer to be 13 and the target layer to be 7.
\subsection{Additional Results of CREME}
We additionally show $\text{IP}(I_o,\widetilde{A_o})$ in Figure~\ref{fig:wrong}. Hopefully, a good correction method has little positive improvement for the prediction of wrong answer $\widetilde{A_o}$. We observe that $p(\widetilde{A_o}|I_o)$ approximately remains unchanged for CREME, while is apparently enlarged with Memory Injection and PatchScopes.
In Figure~\ref{fig:edit_layer}, we show the effects of different editing layer, where the effect of editing layer 19 largely surpasses editing other layers (5,10,23,29). This results align well with the results of the locating experiment (Figure~\ref{fig:locating_dataset}).
\subsection{Showcase of CREME}
We use specific cases to show the effect of leveraging CREME to correct the compositional reasoning failures of LLaMA-2-7B in Table~\ref{tab:showcase1}.
\begin{table}[t]
\centering
\resizebox{0.5\textwidth}{!}{
\begin{tabular}{lcc}
\hline
\textbf{Method} & ROME \footnotesize{(w. ground-truth)} & CREME \footnotesize{(w.o. ground-truth)}   \\
\hline
\textit{Correction}($\uparrow$)& $\mathbf{98.0}\%$ & $95.3\%$  \\
\textit{Paraphrasing}($\uparrow$)&$62.5\%$  & $\textbf{70.5}\%$  \\
\textit{Generalization}($\uparrow$)&$+1.24$  & $+\textbf{3.61}$  \\
\textit{Specificity}($\downarrow$)&$+5.37$  & $+\textbf{1.24}$  \\
\hline 
\end{tabular}
}
\vspace{-0.1in}
\caption{
Comparing CREME and ROME~\cite{meng2022locating} (applied on OpenAlpaca-3B). ``w. ground-truth” refers to that ROME requires $A_o$ for editing.
}
\vspace{-0.1in}
\label{tab:edit_rome}
\end{table}

\begin{table}[t]
\centering
\resizebox{0.5\textwidth}{!}{
\begin{tabular}{lcc}
\hline
\textbf{Input Types} & Paraphrasing $I_p$  & Generalization $I_g$  \\
\hline 
\hline
$\Delta\log\text{PPL}$ & $-29.9\%$ & $-17.9\%$ \\
\hline
\hline
Pred Accuracy \\
\hline
\textit{Pre-Patching} & $6.3\%$ & $29.9\%$  \\
\textit{Post-Patching} & $\mathbf{18.3}\%$ & $\mathbf{38.3}\%$ \\
\hline
\end{tabular}
}
\vspace{-0.1in}
\caption{
$\Delta\log\text{PPL}$ and Prediction Accuracy results for LLaMA-2-7B.
}
\vspace{-0.1in}
\label{fig:edit_final_predict}
\end{table}